\RequirePackage{fix-cm}

\documentclass[twocolumn]{svjour3}          
\smartqed  
\usepackage{graphicx}
\usepackage{natbib}
\usepackage{multirow}
\usepackage{amsmath}
\usepackage{amssymb}
\usepackage{bbm}
\usepackage{bm}
\usepackage{booktabs}
\usepackage{array} 
\usepackage{blindtext}
\usepackage{comment}

\usepackage{algorithm}
\usepackage{algorithmic}

\usepackage{subcaption}

\usepackage{xspace}

\makeatletter
\DeclareRobustCommand\onedot{\futurelet\@let@token\@onedot}
\def\@onedot{\ifx\@let@token.\else.\null\fi\xspace}

\def\ie{\emph{i.e}\onedot} 
 
 \def\vs{\emph{vs}\onedot}

\makeatother

\usepackage{color, colortbl}
\usepackage[table]{xcolor}
\definecolor{remark}{rgb}{1,.5,0} 
\definecolor{citecolor}{rgb}{0,0.443,0.737} 
\definecolor{linkcolor}{rgb}{0.956,0.298,0.235} 
\definecolor{gray}{gray}{0.95}
\definecolor{cyan}{rgb}{0.831,0.901,0.945}
\definecolor{mygray}{gray}{.9}

\definecolor{lightgreen}{HTML}{D8ECD1}
\definecolor{edit}{HTML}{C63678}

\usepackage{pifont}

\usepackage{color, colortbl}
\definecolor{citecolor}{HTML}{0071bc}
\definecolor{tabhighlight}{HTML}{e5e5e5}
\usepackage[colorlinks,citecolor=citecolor]{hyperref}

\makeatletter
\renewcommand\paragraph{
  \@startsection{paragraph} 
  {4} 
  {\z@} 
  {.5em \@plus1ex \@minus.2ex} 
  {-.5em} 
  {\normalfont\normalsize\bfseries} 
}
\makeatother

\begin{document}
\sloppy

\title{SCT: A Simple Baseline for Parameter-Efficient Fine-Tuning via Salient Channels 
}

\titlerunning{Salient Channel Tuning}        

\author{Henry Hengyuan Zhao         \and
        Pichao Wang \and
        Yuyang Zhao \and
        Hao Luo \and
        Fan Wang \and
        Mike Zheng Shou
}


\institute{Henry Hengyuan Zhao \at
              Show Lab, National University of Singapore, Singapore \\
              \email{hengyuan.z@u.nus.edu}
           \and
           Pichao Wang \at
                Alibaba Group, USA \\
              \email{pichaowang@gmail.com}
           \and
           Yuyang Zhao \at
              National University of Singapore, Singapore \\
              \email{yuyang.zhao@u.nus.edu}
           \and
           Hao Luo \at
           Alibaba Group, China \\
           \email{michuan.lh@alibaba-inc.com}
            \and
           Fan Wang \at
           Alibaba Group, USA \\
           \email{fan.w@alibaba-inc.com}
           \and
           Mike Zheng Shou (Corresponding author)\at
           Show Lab, National University of Singapore, Singapore \\
           \email{mike.zheng.shou@gmail.com}
}

\date{Received: date / Accepted: date}

\maketitle

\begin{abstract}

Pre-trained vision transformers have strong representation benefits to various downstream tasks. Recently, many parameter-efficient fine-tuning (PEFT) methods have been proposed, and their experiments demonstrate that tuning only 1\% extra parameters could surpass full fine-tuning in low-data resource scenarios. However, these methods overlook the task-specific information when fine-tuning diverse downstream tasks. In this paper, we propose a simple yet effective method called ``Salient Channel Tuning" (SCT) to leverage the task-specific information by forwarding the model with the task images to select partial channels in a feature map that enables us to tune only 1/8 channels leading to significantly lower parameter costs. Experiments on 19 visual transfer learning downstream tasks demonstrate that our SCT outperforms full fine-tuning on 18 out of 19 tasks by adding only 0.11M parameters of the ViT-B, which is 780$\times$ fewer than its full fine-tuning counterpart. Furthermore, experiments on domain generalization and few-shot classification further demonstrate the effectiveness and generic of our approach. The code is available at \href{https://github.com/showlab/SCT}{https://github.com/showlab/SCT}

\end{abstract}

\section{Introduction}
\label{sec:intro}

\begin{figure}[t]
\begin{center}
\includegraphics[width=\linewidth]{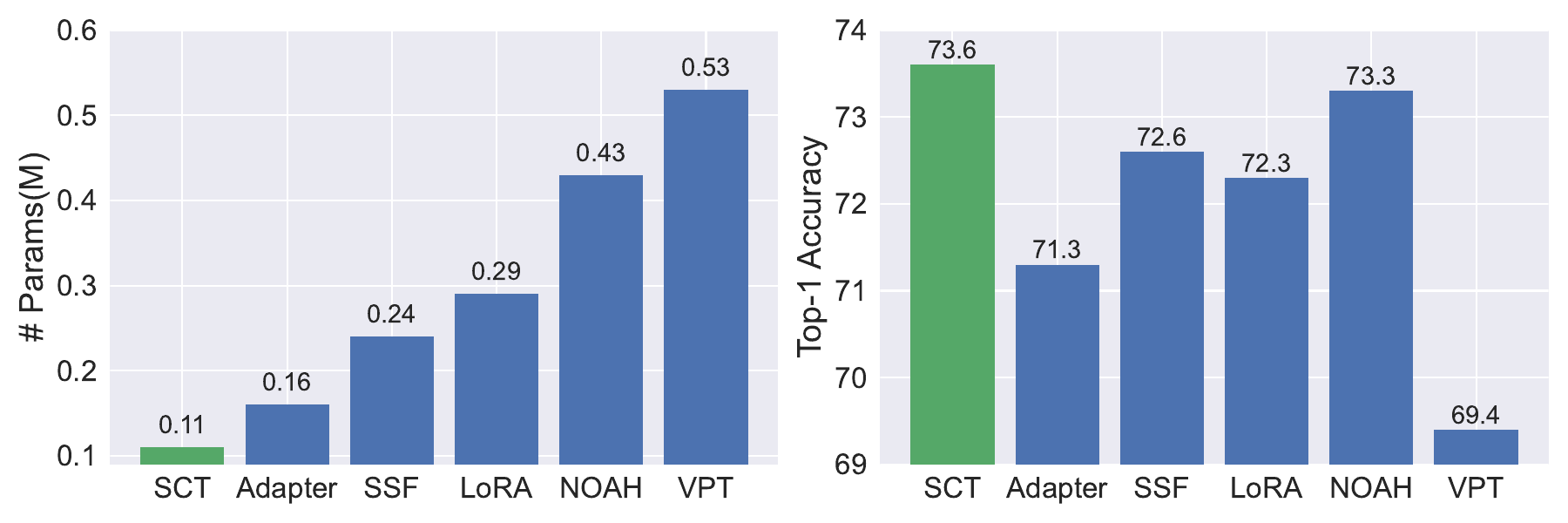}
\end{center}
\caption{The comparison of parameters and top-1 accuracy on VTAB-1K benchmark with different baselines. We only tune 96 channels in 768 channels of ViT-B/16, obtaining the best results compared with other methods.}
\label{fig:teaser}
\end{figure}

Large vision transformers (ViT) have achieved remarkable success in computer vision tasks~\citep{dosovitskiy2020image, liu2021swin, yuan2022volo,zhou2021elsa, carion2020end, li2021benchmarking, strudel2021segmenter} by using large-scale training data, such as ImageNet21K and JFT-300M, to create strong representations. However, downstream recognition tasks often lack sufficient data to train from scratch. Therefore, transferring knowledge from pre-trained ViT models can significantly reduce training difficulty and produce promising results. End-to-end full fine-tuning is a widely used way to inherit these robust representations, but it faces two challenges. First, large models are prone to overfitting when tuning their massive weights on small downstream training data. Second, ViT models are too large to store all weights for each downstream task, making it infeasible to deploy fine-tuned models on resource-limited devices.

\begin{figure}
\begin{center}
\includegraphics[width=0.9\linewidth]{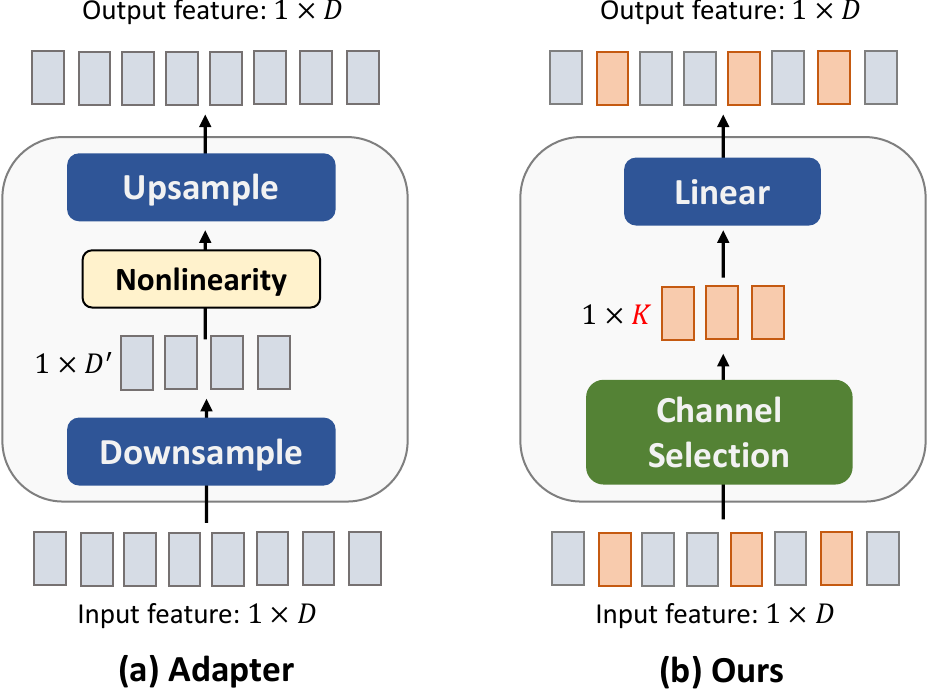}
\end{center}
\caption{The architecture comparison between the Adapter and our SCT. ``Downsample" and ``Upsample" represent the channel downsampling and upsampling operations. $D$ represents the number of channel dimensions.}
\label{fig:compareadapter}
\end{figure}

\begin{figure*}[!ht]
\begin{center}
\includegraphics[width=0.85\linewidth]{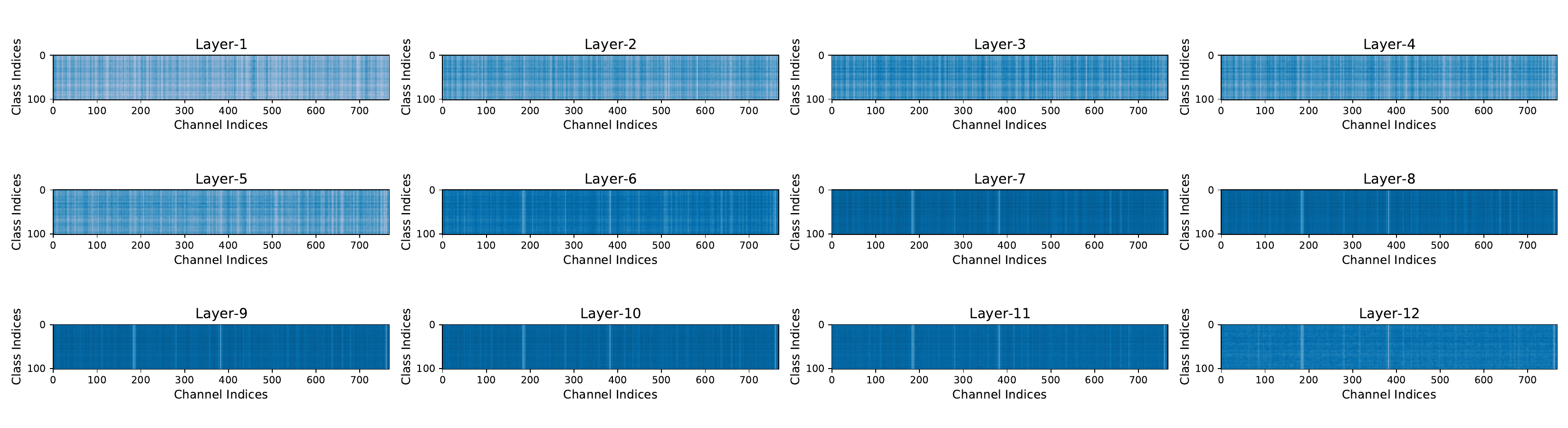}
\subcaption[]{Visualizations of the extracted feature maps on the \textit{Caltech101} dataset at each transformer layer. Y-axis represents the class indices, and X-axis represents channel indices, \ie, 768 in total. We categorize the total image features with the class label and then calculate each channel's $L_2$ values. Thus, we obtain the 768 dimension vector for each class and find that some channels have higher activation values than others in all classes, as the vertical lines are shown in this figure.}
\end{center}
\begin{center}
\includegraphics[width=0.85\linewidth]{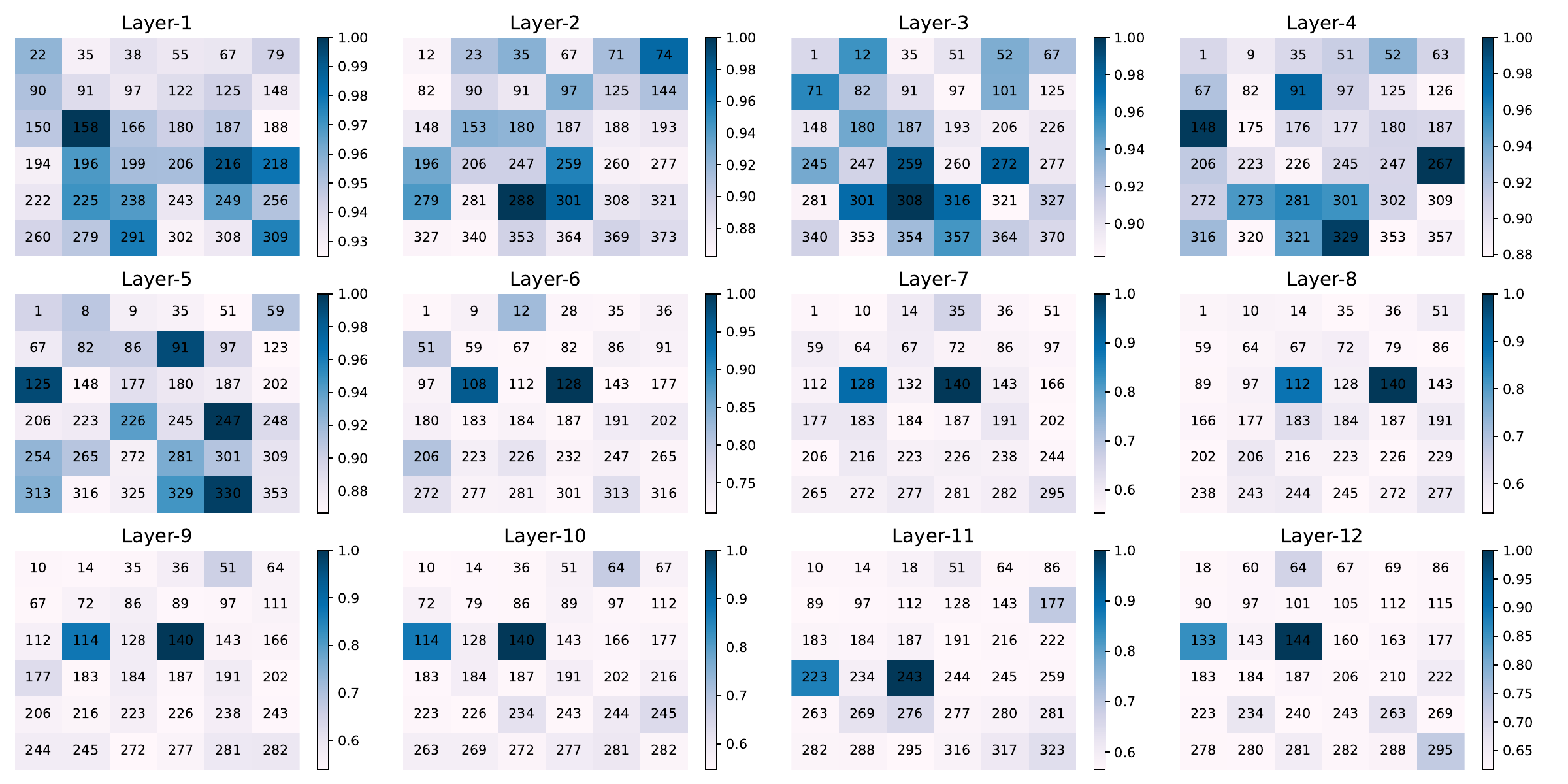}
\subcaption[]{The top-36 selected salient channel indices on the \textit{Caltech101} dataset at each transformer layer \ie, Layer-1, Layer-2. Each layer selects different salient channels. As the layer goes deeper, the salient channels appear more concentrated. Darker color represents a larger activation value.}
\end{center}
\vspace{-0.1in}

\begin{center}
\includegraphics[width=0.85\linewidth]{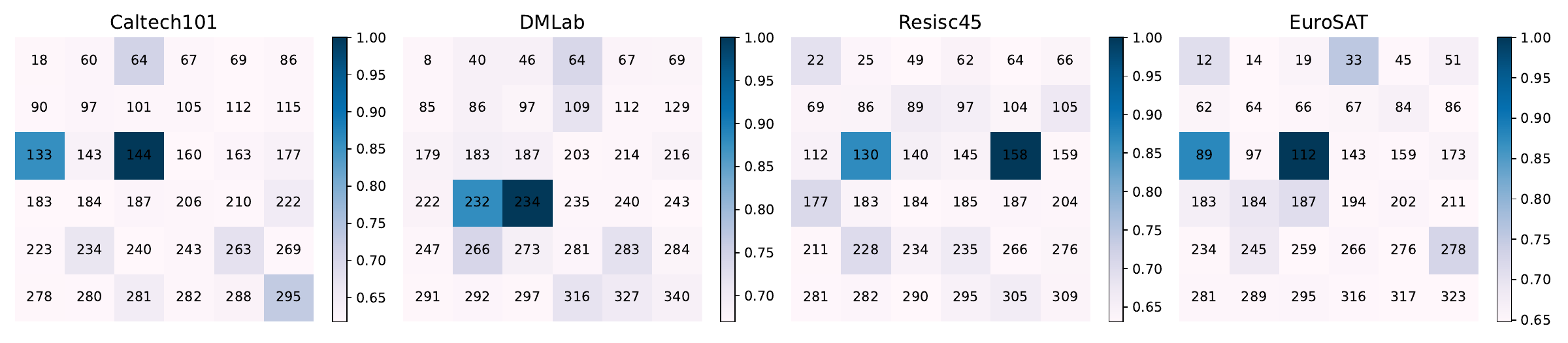}
\subcaption[]{We select the salient channels from the same layer on four downstream tasks. The results suggest that channel bias exists in various tasks.}
\end{center}
\vspace{-0.1in}
\caption{The visualizations of feature maps and the selected salient channel indices. All the results are obtained by ViT-B/16 pre-trained on ImageNet21K.}
\label{fig:fig1}
\vspace{-0.1in}
\end{figure*}

To mitigate the above two challenges, some works propose to tune a subset of parameters \citep{zaken2021bitfit} or adopt an external trainable module \citep{adapter,lora} to preserve the knowledge learned from pre-trained models.
As for tuning a subset of parameters, there are two representative approaches: tuning the classification head \citep{mahajan2018exploring, jia2021exploring, chen2021empirical} and tuning bias term \citep{cai2020tinytl}. 
Tuning the classification head involves freezing the weights of the backbone network and updating only the linear head, whereas tuning the bias term entails unfreezing the bias term within the backbone network. Both approaches result in inferior performance.

Recently, some studies proposed to address a new task named Parameter-Efficient Fine-Tuning (PEFT) by leveraging an external trainable module for model adaptation. One type of method includes integrating learnable prompts into the model input for each downstream task, exemplified by the pioneering work, VPT\citep{vpt}. However, VPT encounters several challenges. First, its task-specific prompts are derived through supervised learning, which differs from the user-provided text prompts commonly used in the NLP field. Second, VPT requires searching for the best prompt length for each task which lacks flexibility and is impractical when encountering new downstream tasks. Another type of work is to add an adapter \citep{chen2022adaptformer, adapter, jie2022convolutional} alongside the multi-head self-attention (MHSA) or MLP block and address the whole features equally without considering the task-specific information into the efficient module design. Moreover, the number of trainable parameters is not quite small.

By investigating these approaches, utilizing the task-specific information in the efficient module design is still underexplored as it is highlighted in a recent study \citep{luo2022channel} that channel bias exists in diverse downstream tasks. In this paper, to address the PEFT problem, we propose a simple baseline to involve task-specific information by forwarding the pre-trained model with downstream images. Furthermore, we proceed to select a small portion of channels to achieve efficient fine-tuning, significantly reducing the parameter costs. Our experiments in Sec.\ref{sec:expvtab1k} reconfirm that tuning only a small portion of task-specific channels is sufficient for downstream task adaptation in the low-data regime. To achieve the goal of obtaining these task-specific channels, we propose a Class-Aware Importance Score (CAIS) by adopting $L_2$ norm as the criteria to evaluate the channel importance inspired by the classic pruning method \citep{han2015learning}. We refer to the selected channels as ``salient channels" (SC) since they have higher activation values and are crucial for task performance. It is noted that channel selection can avoid the training costs by selecting the best structure as in NOAH\citep{noah} or the best prompt length as in VPT\citep{vpt}. 
Moreover, compared with adapter-based methods~\citep{adapter, chen2022adaptformer}, selecting partial task-specific channels enables us to achieve lower parameter costs by removing the operations of ``downsample" and ``nonlinearity" as shown in Fig.~\ref{fig:compareadapter}.

In summary, the contributions are summarized as follows:

\noindent\textbf{Contributions}

\begin{itemize}
    \item We propose a simple baseline with a new perspective in partial channel tuning (\ie, salient channel tuning) for addressing the PEFT task to leverage task-specific information, achieving promising performance on 19 visual transfer learning tasks, 4 domain generalization tasks, and 5 few-shot learning tasks. The experiments demonstrate that tuning a subset of channels in a feature map is efficient and generic.
    \item To the best of our knowledge, it is the first time to reduce the parameter costs to 0.11M (780$\times$ fewer parameters than the full model size) level of ViT-B while achieving the best average performance compared with other methods.
\end{itemize}

\section{Related Work}
\raggedbottom

\subsection{Vision Transformers}

Transformer \citep{vaswani2017attention} has demonstrated outstanding results on natural language processing and computer vision tasks. Lots of vision transformers \citep{chen2021crossvit,d2021convit,dong2022cswin,ali2021xcit,fan2021multiscale,han2021transformer,rao2021dynamicvit,yuan2021tokens,touvron2021going,liu2021swin,wang2021kvt,zhou2021elsa} are proposed after the pioneering work ViT \citep{dosovitskiy2020image}. Many of them increase the model size gradually for state-of-the-art results and learn the rich representations by various architectural designs. It is noted that most of them are trained on the natural dataset and have the strong potential to be transferred to other domains/tasks. Moreover, adopting these models to the downstream tasks is able to alleviate the training difficulty and achieve promising results. 

\subsection{Parameter-Efficient Fine-Tuning Methods}

PEFT focuses on adopting a trainable module with a few parameters for fine-tuning. 
Two lines of PEFT have been proposed recently. On the one hand, applying prompts \citep{vpt,liu2022prompt,xing2022class,zheng2022prompt,nie2022pro,wang2022fine,zhou2022cocoop, zhou2022coop,liao2023rethinking,shu2022tpt,zhang2023VisualPromptRetrieval,zang2022unified,visprompt} to the backbone networks shows success on several vision tasks. VPT~\citep{vpt} first proposes the prompt-based method in computer vision fields by injecting the prompts into each transformer layer. However, one main limitation of VPT is that it relies on hand-crafted selection to determine the optimal prompt length for each task. This is inflexible when applying to a new downstream task. Another limitation is that the task-specific prompts of VPT are obtained through the training procedure and cannot be given by the user as in the NLP field. In our work, to explicitly reduce the computations of searching the task-specific prompt length, we pass the training images into the backbone network and leverage the task-specific information by determining the salient channels. This procedure only needs one forward propagation without training costs.

On the other hand, adding a residual module \citep{adapter, chen2022adaptformer, jie2022convolutional,chen2022conv} in the backbone networks also acquires promising results for performance and efficiency.
Adapter \citep{adapter}, a widely used baseline in many tasks ~\citep{sung2022vl, pan2022st, zhang2023multimodal}, proposes an MLP-like module, a successful design that first projects the original dimensional features into a smaller dimension with one nonlinear layer and projects it back to the original dimensions. It vastly reduces the number of parameters. Inspired by its smallest intermediate dimensions, finding a small number of salient channels in a feature map might be enough for the adaptation. Unlike injecting trainable modules into the transformer blocks, LoRA \citep{lora} optimizes a low-rank decomposition matrix with a low intrinsic dimension to project the query, key features. As for NOAH \citep{noah}, a neural architecture search algorithm incorporates Adapter, LoRA, and VPT into its network search space. NOAH provides a strong baseline for performing consistently well on different datasets. SSF~\citep{SSF}, a recently proposed strong baseline by only scaling and shifting the features to implement the efficient model tuning. Child-Tuning~\cite{xu-etal-2021-childtuning} updates a subset of parameters of large pre-trained language models via strategically masking out the gradients of the non-child network during the backward process. AdaptFormer \citet{chen2022adaptformer} is an adapter-like method. It explores the efficiency of the adapter-like architecture of vision transformers' fine-tuning.

Unlike the above methods, our SCT handles the PEFT by leveraging the task-specific information into the method design and then proposes the Salient Channel Tuning, achieving lower parameter costs and higher performance than previous methods.

\section{Method}

\raggedbottom

\begin{figure*}[t]
\begin{center}
\includegraphics[width=1\linewidth]{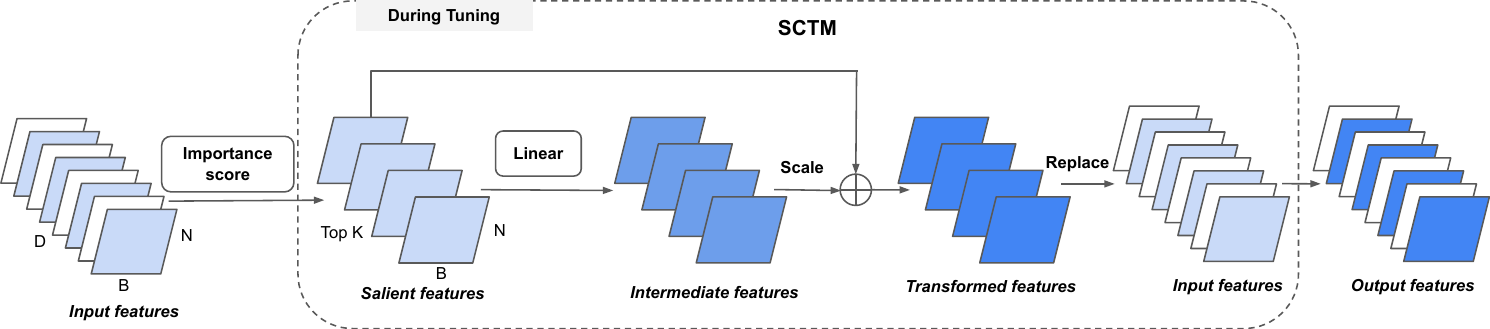}
\end{center}
\caption{The overview of proposed salient channel tuning module.}
\label{fig:overview-ict}
\end{figure*}

\subsection{Not all channels are equal}

To leverage task-specific information, this paper aims to find task-specific channels for fine-tuning. Here we provide a simple observation highlighting the ``Salient Channels" that exist in the downstream tasks.

Previous works \citep{luo2017thinet,li2016pruning, he2018soft, liu2018rethinking, han2015learning, li2017pruning} demonstrate that pruning some channels of deep neural networks has a marginal influence on the model performance but can significantly reduce the parameter number and computational cost. Such results reflect that the importance of different channels is not the same, \ie, ``Not all channels are equal". Intuitively, the channel importance is different in terms of the tasks, which motivates us to investigate the impact of channel selection in model tuning.

To find task-specific channels, an intuitive way is to find some mutual channels across different categories. Thus, We first illustrate an observation between the pre-trained model and the downstream task. We choose \textit{Caltech101} \citep{fei2004learning} (one of the downstream tasks from the VTAB-1K benchmark) as an example to evaluate the intermediate features of each transformer layer of ViT. ViT-B pre-trained on ImageNet21K is the backbone, and we feed the whole dataset to extract the features between multi-head self-attention (``Attn") and ``MLP" blocks in all 12 transformer layers. 
Fig. \ref{fig:fig1} (a) shows some vertical lines in each subfigure of transformer layers.
It indicates that when using a pre-trained model to extract the features on the target dataset, some channels have higher activation values than others, regardless of categories. We so-call these channels ``salient channels". To investigate the position of salient channels, we also report the index of selected top-36 salient channels after calculating the importance score of each layer in Fig. \ref{fig:fig1} (b). The number represents the channel index of the total of 768 channels. The darker color represents the higher activation values. We could find different layers have different salient channel sets. As the layer goes deeper, the salient channels appear more concentrated. We hypothesize that deeper layers contain more abstract information and information aggregated in a few channels. In contrast, the shallow layers extract more low-level representation, making the information dense.

It naturally raises a question: \textbf{Could we find salient channels in a feature map and then only tune these salient channels for efficient tuning?} To answer this question, we design a simple class-aware importance score to identify these salient channels and leverage a salient channel tuning module for efficient fine-tuning.

\subsection{Class-Aware Importance Score}
\label{sec:class-aware}
The selection of salient channels is crucial for the adaptation to downstream tasks, which should contain mutual information in all classes. Intuitively, we can directly use the mean feature of all the data samples to represent the downstream task distribution to determine the salient channels. However, considering downstream datasets may not always be balanced, treating the whole dataset as a unity may be biased to the head classes, leading to the selected channels not representing the mutual information over all the classes. 
Consequently, we propose the Class-Aware Importance Score (CAIS), where the importance score calculation is conducted at each class and then averaged across all the classes.
Assume the intermediate feature maps as $\{f_{m}^{l} \in \mathbb{R} ^{B_{m} \times N \times D} | \ l, m \in \mathbb{N}, 1 \leq l \leq L \ and \  1 \leq m \leq M \}$, where the $N$ and $D$ represent the number of tokens and amount of channel dimension, respectively. $L$ represents the total layers of the ViT backbone. The $B_{m}$ is the volume of each class, and $M$ represents the number of categories of the downstream recognition task. We first apply the $L_2$ norm regularization of each channel dimension of feature $f_{m}^{l}$:
\begin{equation}
\begin{split}
\tilde{f}_{m}^l = & \{{\| f_{m,1}^{l} \|_2},\ {\| f_{m,i}^{l} \|_2},...,\ {\| f_{m,D}^{l} \|_2} \}, \\
& \ \| f_{m,i}^{l} \|_2 \in \mathbb{R},\ i \in \mathbb{N},\ 1 \leq i \leq D,
\end{split} 
\label{eqn:eq1}
\end{equation}
where $\tilde{f}_{m}^l \in \mathbb{R} ^{1 \times D}$ is the importance score vector at $l$-th layer and $m$-th class. To investigate the salient channels, we average the $\tilde{f}_{m}^l \in \mathbb{R} ^{1 \times D}$ over all classes to get the importance score at $l$-th layer:
\begin{equation}
        Z^l = \frac{1}{M} \sum_{m=1}^{M} \tilde{f}_{m}^l,
        \tilde{f}_{m}^l \in \mathbb{R}^{1 \times D},
    \label{eqn:eq2}
\end{equation}

After getting the importance score vector $\{ Z^l \in \mathbb{R} ^{1 \times D} |\ l \in \mathbb{N}, \ 1 \leq l \leq L \}$, we can choose the largest $K$ values of $Z^l$ and then we can derive selected indices as $I^l = top K(Z^l),\ I^l \in \mathbb{N} ^{1 \times K}$ at $l$-th layer. Later, the salient channels at each layer are different across different downstream tasks, and it could explicitly involve task-specific information in the model fine-tuning. The procedures for selecting salient channels are shown in Algorithm \ref{alg:overall}.

\subsection{Adapting ViT via Salient Channel Tuning}
Given the importance score of selecting top-$K$ salient channels, we propose a Salient Channel Tuning Module (SCTM) in this paper. The overview of the SCTM is depicted in Fig. \ref{fig:overview-ict}.
Unlike other PEFT methods, our SCTM only contains a linear layer rather than an MLP-like adapter \citep{adapter} involving two learnable layers and one activation layer illustrated in Fig.\ref{fig:compareadapter}. Such a simple structure is easy to reproduce and maintains relatively low parameter costs. To inherit the original robust representations, we utilize a residual shortcut to fuse the salient features with intermediate features and use the $Scale \in \mathbb{R}$ (as shown in Fig. \ref{fig:overview-ict}) to adjust the weights between both features, which is a constant parameter. After obtaining the transformed features, we insert the transformed features back into the input features at the same position. The unselected channels we keep frozen during fine-tuning. In Fig. \ref{fig:insertposition}, we present two forms of inserting the SCTM into the ViT. The ``Ours-MLP" represents that we insert the SCTM after the MLP block, and the ``Our-Attn" indicates that we put the SCTM after the MHSA block while before the MLP block. In the following experiments, ``Ours-Attn" is the default injection position compared with other baselines.

\begin{figure*}[t]
\begin{center}
\includegraphics[width=0.95\linewidth]{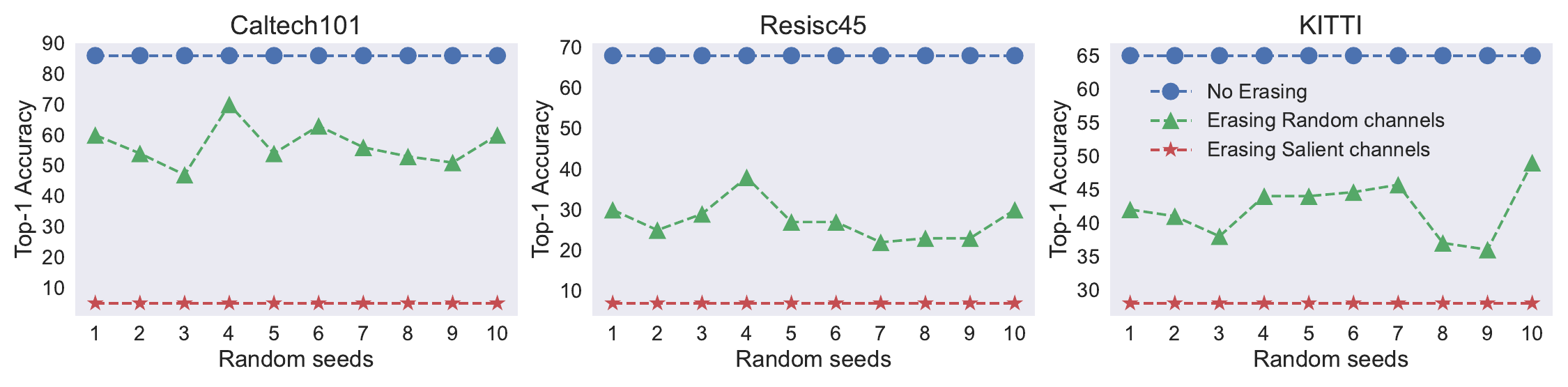}
\subcaption{Evaluate the importance of salient channels on three datasets. We erase all layers with randomly selected channel indices and salient channel indices. To avoid the noise, we repeat the random selection process 10 times. The results demonstrate the effectiveness of our salient channel selection.}
\end{center}
\vspace{-0.2in}


\begin{center}
\includegraphics[width=0.95\linewidth]{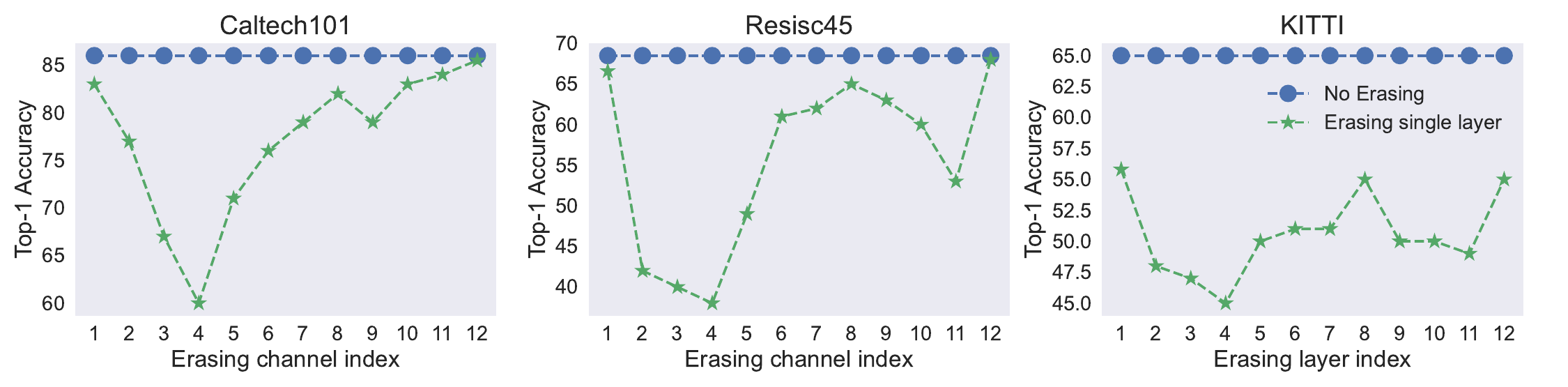}
\subcaption{We systematically remove one layer at a time, identified by its index. The resulting curve highlights the relative significance of the first four layers compared to the last four in the context of the Caltech101 and Resisc45 tasks. When considering the KITTI task, all layers are important for the final output.}
\end{center}
\vspace{-0.2in}

\caption{Channel erasing test.}

\label{fig:channelera}
\end{figure*}

\noindent\textbf{Overall.} Our SCT first feeds the training set to a backbone network to extract the intermediate features in different layers. Then, using the proposed CAIS to determine the salient channels and save the channel indices. During fine-tuning, we add our SCTM in each transformer layer and tune the selected channels with a linear layer while freezing other unselected channels.

\noindent\textbf{Discussion.} 
Firstly, different downstream tasks have their peculiarities, \ie, ``Not all channels are equal". 
Previous PEFT methods (\ie, Adapter) lack the consideration of explicitly using task-specific information. Instead, we propose the SCT explicitly leverage the task-specific information for channel selection, which could enable us to save the computations and extra parameters of the ``Downsample" and ``Nonlinearity" operations compared with Adapter~\citep{adapter}.
Secondly, calculating the salient channels offline could save the computations of searching the best prompt length~\citep{vpt} or the best network structure~\citep{noah}. Thirdly, our results in Fig.~\ref{fig:teaser} demonstrate that only adapting 12.5\% channels could obtain competitive results compared to other baselines. Simultaneously, transforming a small portion of the features could significantly reduce the number of learnable parameters (e.g., $12 \times 96 \times 96$ is relatively more minor than the $12 \times 768 \times 768$). Fourthly, such a small amount of parameters largely reduces the difficulty of storing task-specific knowledge when we have many downstream tasks.
To evaluate this simple baseline, we also test its few-shot capacity and domain generalization ability to meet the requirements of real-world applications.

\subsection{Channel Erasing Test}
To assess the channel selection configuration, we give two toy tests to evaluate the effectiveness of salient channels as in Figure \ref{fig:channelera}. The first test in Figure \ref{fig:channelera} (a) demonstrates the effectiveness of salient channels compared with random channel selection. The second test in Figure \ref{fig:channelera} (b) evaluates the importance of each layer. We erase the salient channels layer by layer. The curves suggest that for Natural type task Caltech101, the first four layers are more important than the last four layers as the accuracy drops significantly in the erasing first four layers. A similar trend is evident in the Resisc45 task. As for the KITTI task, the last four layers are as important as the first four.

Based on the aforementioned tests, we manipulate the number of fine-tuned channels within the first four layers and the last eight layers, as outlined in Table \ref{tab:layernumber}. Increasing the number of fine-tuned channels within the first four layers (first and third rows) yielded enhancements of 0.1\% for Caltech101 and 0.5\% for KITTI, albeit with a 0.1\% decline in Resisc45 performance. Notably, the influence of the last four layers on the final results is slight. Consequently, we explored the effect of decreasing the number of tuned channels within the last eight layers.
The results (fourth and fifth rows) detailed in Table \ref{tab:layernumber} indicate that reducing the $K$ from 96 to 32 leads to a significant deterioration in performance. However, a slightly higher value compared with setting $K=32$ across all 12 layers.

\begin{algorithm}[t]
    \caption{The procedures of channel selection via class-aware importance score.}
    \label{alg:overall}
    \textbf{Inputs:}
    
    The pre-trained ViT model $\mathcal{F}$ and the number of layers $L$; \\
    The number of selected channels $K$ and the total number of classes $M$; \\
    \vspace{-.15in}
    \begin{algorithmic}[1]
        \FOR{each $l \in [1, L]$}
  \STATE Extract the feature $f_{m}^{l}$ at each class $m$;
  \STATE Apply the $L_2$ norm regularization on each channel to obtain $\tilde{f}_{m}^l$ via Eq.~\ref{eqn:eq1};
  \STATE Calculate the importance score $Z^l$ of $l$-th layer via Eq.~\ref{eqn:eq2};
  \STATE Select top-$K$ largest channels $I^l = \text{topK}(Z^l)$;
        \ENDFOR
    \end{algorithmic}
    \textbf{Outputs:} Selected channel indices of each transformer layer.
\end{algorithm}

\begin{table*}[]
\begin{center}
\resizebox{0.8\linewidth}{!}{
\begin{tabular}{lcccc}
\toprule
 Channel configurations& Caltech101 & Resisc45 & KITTI & Params\\
\midrule
\multirow{1}{*}{[96, 96, 96, 96, 96, 96, 96, 96, 96, 96, 96, 96]}  & 91.6 & 86.3 & 80.2 & 0.11M \\
\multirow{1}{*}{[192, 192, 192, 192, 192, 192, 192, 192, 192, 192, 192, 192]}  & 91.5 & \textbf{87.0} & \textbf{81.0} & 0.44M\\
\multirow{1}{*}{[192, 192, 192, 192, 96, 96, 96, 96, 96, 96, 96, 96]}  & \textbf{91.7} & 86.2 &  80.7 & 0.22M\\
\multirow{1}{*}{[96, 96, 96, 96, 32, 32, 32, 32, 32, 32, 32, 32]}  & 91.1 & 85.0 & 76.2 & 0.045M \\

\multirow{1}{*}{[32, 32, 32, 32, 32, 32, 32, 32, 32, 32, 32, 32]}  & 90.3 & 84.8 & 74.0 & 0.01M \\
\bottomrule
\end{tabular}
}
\end{center}
\vspace{-0.1in}
\caption{Different number of fine-tuned channels on three tasks.}
\label{tab:layernumber}
\end{table*}

\section{Experiments}

This section compares our SCT with other state-of-the-art PEFT baselines on the VTAB-1K benchmark, using ViT and Swin Transformer backbones. In addition, we analyze the channel selection strategy, class-aware importance score, insert position, insert length, and the number of selected channels to verify SCT's effectiveness further. Beyond evaluating performance on visual transfer learning tasks, we also validate the capabilities of four domain generalization datasets and test the performance of SCT in the low-data scenarios on five datasets in the few-shot setting.

\subsection{Training details}
Following VPT \citep{vpt}, we utilize grid search to find the tuning-specific hyper-parameters, including learning rate and weight decay, based on the val set of each task as shown in Tab.\ref{tab:traindetails}. For the hyper-parameter $Scale$, which is a manually defined constant to balance the weight of transformed features and salient features, we search it at the range \{0.1, 0.2, 0.3, 0.4, 0.5, 0.6, 0.7, 0.8, 0.9, 1.0\}. During training, we use a standard image augmentation strategy: normalize with ImageNet means and standard deviation, randomly resize crop to 224×224, and random horizontal flip as in VPT.

\begin{table*}[ht]
\begin{center}
\resizebox{0.8\linewidth}{!}{
\begin{tabular}{lc}
\toprule
\multirow{1}{*}{} & VTAB-1K, Domain Generalization, and Few-shot Learning\\
\midrule
\multirow{1}{*}{Optimizer} & AdamW \citep{loshchilov2017decoupled}\\
\multirow{1}{*}{$base_{lr}$ range} & \{0.1, 0.5, 0.01, 0.05 0.001, 0.005, 0.0001, 0.0005\}\\
\multirow{1}{*}{Weight decay range} & \{0.1 0.01, 0.05, 0.001, 0.005, 0.0005, 0.0001\}\\
\multirow{1}{*}{Learning rate schedule} & cosine decay\\
\multirow{1}{*}{Warm up epochs} & 10\\
\multirow{1}{*}{Total epochs} & 100\\
\multirow{1}{*}{Batch size} & 64\\

\bottomrule
\end{tabular}
}
\end{center}
\caption{Training details on each visual task with ViT-B/16.}
\label{tab:traindetails}
\end{table*}

\subsection{Experiments on VTAB-1K Benchmark}
\label{sec:expvtab1k}

\begin{table*}[ht]
\begin{center}
\resizebox{0.85\linewidth}{!}{
\begin{tabular}{llcccc}
\toprule
 & Dataset & \# Classes & Train & Val & Test \\
\midrule
\multicolumn{6}{c}{ \textbf{VTAB-1K} \citep{zhai2019large}}\\
\midrule
\multirow{7}{*}{Natural} & CIFAR100 \citep{krizhevsky2009learning} & 100 & \multirow{7}{*}{800/1000} & \multirow{7}{*}{200} & 10,000 \\
& \multirow{1}{*}{Caltech101 \citep{fei2004learning}} & 102 &  &  &  6,048\\
& \multirow{1}{*}{DTD \citep{cimpoi14describing}} & 47 &  &  &  1,880\\
& \multirow{1}{*}{Oxford-Flowers102 \citep{nilsback2006visual}} & 192 &  &  & 6,149 \\
& \multirow{1}{*}{Oxford-PetS \citep{parkhi2012cats}} & 37 &  &  & 3,669 \\
& \multirow{1}{*}{SVHN \citep{netzer2011reading}} & 10 &  &  & 26,032 \\
& \multirow{1}{*}{Sun397 \citep{xiao2010sun}} & 397 &  &  & 21,750 \\
\midrule
\multirow{4}{*}{Specialized}& Patch Camelyon \citep{Veeling2018qh} & 2 & \multirow{4}{*}{800/1000} & \multirow{4}{*}{200} & 32,768 \\
& \multirow{1}{*}{EuroSAT \citep{helber2019eurosat}} & 10 &  &  &  5,400\\
& \multirow{1}{*}{Resisc45 \citep{cheng2017remote}} & 45 &  &  &  1,880\\
& \multirow{1}{*}{Retinopathy \citep{kaggle2015retinopathy}} & 5 &  &  & 42,670 \\
\midrule
\multirow{8}{*}{Structured}& Clevr/count \citep{johnson2017clevr} & 8 & \multirow{8}{*}{800/1000} & \multirow{8}{*}{200} & 15,000 \\
& \multirow{1}{*}{Clevr/distance \citep{johnson2017clevr}} & 6 &  &  &  15,000\\
& \multirow{1}{*}{DMLab \citep{beattie2016deepmind}} & 6 &  &  &  22,735\\
& \multirow{1}{*}{KITTI-Dist \citep{geiger2013vision}} & 4 &  &  &  711\\
& \multirow{1}{*}{dSprites/location \citep{matthey2017dsprites}} & 16 &  &  &  73,728\\
& \multirow{1}{*}{dSprites/orientation \citep{matthey2017dsprites}} & 16 &  &  &  73,728\\
& \multirow{1}{*}{SmallNORB/azimuth \citep{lecun2004learning}} & 18 &  &  &  12,150\\
& \multirow{1}{*}{SmallNORB/elevation \citep{lecun2004learning}} & 18 &  &  &  12,150\\
\midrule
\multicolumn{6}{c}{ \textbf{Domain generalization}}\\
\midrule
& \multirow{1}{*}{ImageNet-1K \citep{deng2009imagenet}} & 1,000 & 16 per class  & 50,000 &  N/A\\
& \multirow{1}{*}{ImageNet-V2 \citep{recht2019imagenet}} & 1,000 & N/A  & N/A &  10,000\\
& \multirow{1}{*}{ImageNet-Sketch \citep{wang2019learning}} & 1,000 & N/A & N/A & 50,889\\
& \multirow{1}{*}{ImageNet-A \citep{hendrycks2021natural}} & 200 & N/A & N/A & 7,500\\
& \multirow{1}{*}{ImageNet-R \citep{hendrycks2021many}} & 200 & N/A & N/A & 30,000\\
\midrule
\multicolumn{6}{c}{ \textbf{Few-shot Learning }}\\
& \multirow{1}{*}{Food101 \citep{bossard2014food}} & 101 & \multirow{5}{*}{1/2/4/8/16 per class}  & 20,200 &  30,300\\
& \multirow{1}{*}{Stanford Cars \citep{krause20133d}} & 196 &  & 1,635 &  8,041\\
& \multirow{1}{*}{Oxford-Flowers102 \citep{nilsback2006visual}} & 102 & & 1,633 & 2,463\\
& \multirow{1}{*}{FGVC-Aircraft \citep{maji2013fine}} & 100 & & 3,333 & 3,333\\
& \multirow{1}{*}{Oxford-Pets \citep{parkhi2012cats}} & 37 & & 736 & 3,669\\

\bottomrule
\end{tabular}
}
\end{center}
\caption{Specifications of used datasets on VTAB-1K benchmark, Domain Generalization, and Few-shot Learning.}
\label{tab:datasets-specifications}
\end{table*}

\noindent\textbf{Dataset.}
VTAB-1K~\citep{zhai2019large} contains 19 visual classification tasks which cover a broad spectrum of domains and semantics in three groups, \ie, \textit{Natural}, \textit{Specialized}, and \textit{Structured}. The \textit{Natural} group contains 7 classic classification datasets of natural images. The \textit{Specialized} group involves 4 datasets of two special scenarios: medical and remote-sensing. The \textit{Structured} group has 8 datasets, mainly focusing on understanding the structure of a scene, such as object counting and depth prediction. Each task of VTAB-1K contains 1000 training images. More details are available in Tab.\ref{tab:datasets-specifications}.
Following \cite{noah, vpt}, we use the 800-200 \textsc{train-val} split to determine the hyperparameters and the entire 1000 training data to train the final model. 
We report the average top-1 accuracy on the \textsc{test} set.

\begin{figure}
\begin{center}
\includegraphics[width=0.75\linewidth]{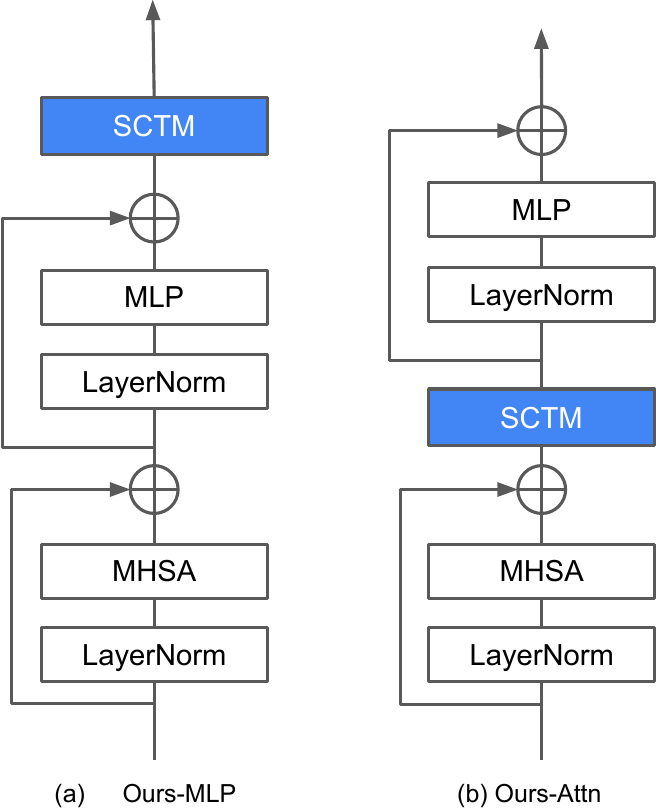}
\end{center}
\caption{Two types of structures when inserting SCTM into the backbone. Only SCTM is trainable, and other modules are frozen.}
\label{fig:insertposition}
\end{figure}

\noindent\textbf{Baselines.}
We compare our method with three baselines \textbf{Full fine-tuning}, \textbf{Linear}, and \textbf{Bias} without external parameters and four baselines \textbf{Adapter}~\citep{adapter}, \textbf{LoRA}~\citep{lora}, \textbf{VPT}~\citep{vpt}, and \textbf{NOAH}~\citep{noah} with external parameters. \textbf{Bias} method only updates all the bias terms in the pre-trained backbone. \textbf{Adapter} injects an additional MLP module into each transformer layer. \textbf{LoRA} adopts an optimized low-rank matrix to the MHSA module in the transformer layers. \textbf{VPT} is a visual prompt algorithm to incorporate the prompts with tokens into the backbone. \textbf{NOAH} is a neural architecture search algorithm incorporating the Adapter, LoRA, and VPT into the network search. \textbf{SSF} propose two learnable factors to scale and shift the features when fine-tuning the downstream tasks. \textbf{AdaptFormer} proposes an adapter-like architecture for vision tasks.
We directly use their released results or run their code to generate them to provide a fair comparison.

\begin{table*}[t]
\begin{center}
\large
\resizebox{\linewidth}{!}{
\begin{tabular}{lc|ccccccc|cccc|cccccccc|c}
\toprule
\multirow{2}{*}{} &  & \multicolumn{7}{c}{\textbf{Natural}} & \multicolumn{4}{c}{\textbf{Specialized}} & \multicolumn{8}{c}{\textbf{Structured}} \\
& \rotatebox{90}{\# Params (M)} & \rotatebox{90}{Cifar100} & \rotatebox{90}{Caltech101}  & \rotatebox{90}{DTD} & \rotatebox{90}{Flower102} & \rotatebox{90}{Pets} & \rotatebox{90}{SVHN} & \rotatebox{90}{Sun397} & \rotatebox{90}{Camelyon} & \rotatebox{90}{EuroSAT} & \rotatebox{90}{Resisc45} & \rotatebox{90}{Retinopathy} & \rotatebox{90}{Clevr-Count} & \rotatebox{90}{Clevr-Dist} & \rotatebox{90}{DMLab} & \rotatebox{90}{KITTI-Dist} & \rotatebox{90}{dSpr-Loc} & \rotatebox{90}{dSpr-Ori} & \rotatebox{90}{sNORB-Azim} & \rotatebox{90}{sNORB-Ele} & \rotatebox{90}{\textbf{Average}}\\
\midrule
\multicolumn{22}{c}{Commonly used methods} \\
\midrule
\multirow{1}{*}{Full} & 85.8 & 68.9 & 87.7 & 64.3 & 97.2 & 86.9 & 87.4 & 38.8 & 79.7 & 95.7 & 84.2 & 73.9 &  56.3 & 58.6 & 41.7 & 65.5 & 57.5 & 46.7 & 25.7 & 29.1 & 65.6\\
\multirow{1}{*}{Linear} & 0 & 64.4 & 85.0 & 63.2 & 97.0 & 86.3 & 36.6 & 51.0 &  78.5 & 87.5 & 68.5 & 74.0 & 34.3 & 30.6 & 33.2 & 55.4 & 12.5 & 20.0 & 9.6 & 19.2 & 53.0\\
\multirow{1}{*}{Bias} & 0.10 & 72.8&87.0&59.2&97.5&85.3&59.9&51.4&78.7&91.6&72.9&69.8&61.5&55.6&32.4&55.9&66.6&40.0&15.7&25.1&62.1\\
\midrule
\multicolumn{22}{c}{PEFT methods} \\
\midrule

\multirow{1}{*}{VPT-Deep} & 0.53 & \textbf{78.8}&90.8&65.8&98.0&88.3&78.1&49.6&81.8&\textbf{96.1}&83.4&68.4&68.5&60.0&46.5&72.8&73.6&\underline{47.9}&\underline{32.9}&37.8&69.4 \\

\multirow{1}{*}{NOAH} & 0.43 & 69.6&\textbf{92.7}&70.2&\underline{99.1}&90.4&86.1&53.7&84.4&95.4&83.9&75.8&\underline{82.8}&\underline{68.9}&49.9&\textbf{81.7}&\textbf{81.8}&\textbf{48.3}&32.8&\underline{44.2}&\underline{73.3}\\

\multirow{1}{*}{LoRA} & 0.29 & 67.1&91.4&69.4&98.8&90.4&85.3&54.0&\underline{84.9}&95.3&84.4&73.6&\textbf{82.9}&\textbf{69.2}&49.8&78.5&75.7&47.1&31.0&44.0&72.3 \\

\multirow{1}{*}{Adapter-8$^{\dagger}$} & 0.16 &72.8&90.7&70.3&93.8&\textbf{91.1}&87.5&\underline{54.8}&83.0&84.9&85.7&75.5&82.1&63.4&50.9&79.2&74.8&47.3&28.6&39.1&71.3\\

\multirow{1}{*}{AdaptFormer-8$^{\dagger}$} & 0.18 & 73.5&91.5&\underline{70.8}&98.9&\textbf{91.1}&\underline{87.8}&54.3&82.3&94.9&\textbf{86.9}&\textbf{76.3}&82.3&66.3&\underline{51.0}&78.8&\underline{76.5}&46.2&32.3&40.1&72.7\\

\midrule

\rowcolor{lightgreen}
\multirow{1}{*}{SCT} & \textbf{0.11} &\underline{75.3}&\underline{91.6}&\textbf{72.2}&\textbf{99.2}&\textbf{91.1}&\textbf{91.2}&\textbf{55.0}&\textbf{85.0}&\textbf{96.1}&\underline{86.3}&\underline{76.2}&81.5&65.1&\textbf{51.7}&\underline{80.2}&75.4&46.2&\textbf{33.2}&\textbf{45.7}&\textbf{73.6}\\

\bottomrule
\end{tabular}
}
\end{center}
\vspace{-.1in}
\caption{Comparisons with state-of-the-art methods on the VTAB-1K benchmark with ViT-B/16. Average results are calculated across 19 datasets. ``\# Params" denotes the backbone's average number of extra trainable parameters. The best performance and smallest parameter number are \textbf{bolded} in each column. \underline{Underline} notes the second-best result. $\dagger$ means the model we implemented and trained on the same training setting, and the hyperparameters are selected by standard grid search since Adapter \cite{adapter} is a strong baseline for NLP tasks and Adaptformer \cite{chen2022adaptformer} is not evaluated on VTAB-1K. The results of LoRA \cite{lora} are from NOAH \cite{noah}. The hidden dimensions of the Adapter and AdaptFormer are 8 and 96. The reduction dimension of LoRA we set is 8.}

\label{tab:vit-vtab-full}

\end{table*}

\begin{figure*}[ht]
\begin{center}
\includegraphics[width=\linewidth]{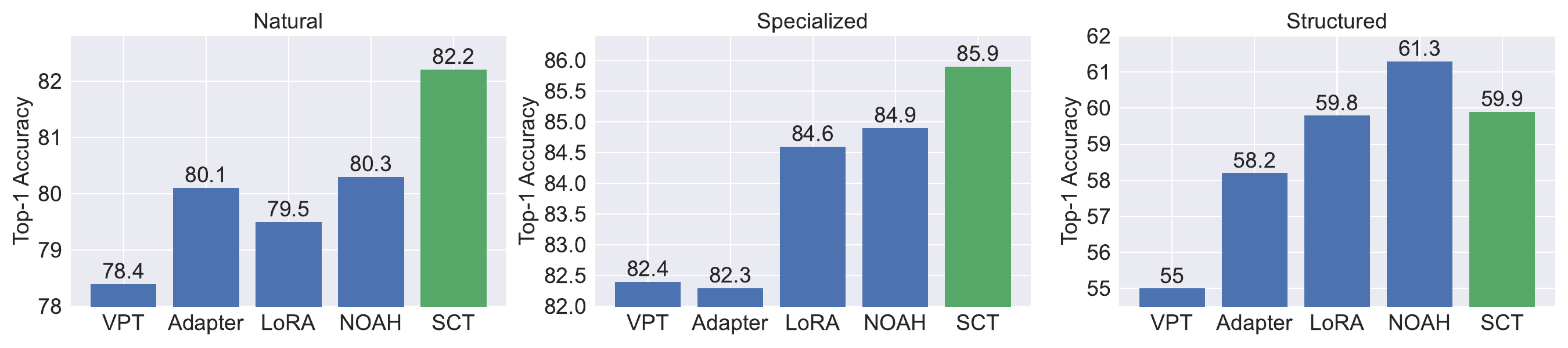}
\end{center}
\caption{Group-wise average results on VTAB-1K benchmark.}
\label{fig:maingroup}
\end{figure*}

\begin{table*}[t]
\begin{center}
\resizebox{\linewidth}{!}{
\begin{tabular}{lc|ccccccc|cccc|cccccccc|c}
\toprule
\multirow{2}{*}{} &  & \multicolumn{7}{c}{\textbf{Natural}} & \multicolumn{4}{c}{\textbf{Specialized}} & \multicolumn{8}{c}{\textbf{Structured}} \\
& \rotatebox{90}{\# Params (M)} & \rotatebox{90}{Cifar100} & \rotatebox{90}{Caltech101}  & \rotatebox{90}{DTD} & \rotatebox{90}{Flower102} & \rotatebox{90}{Pets} & \rotatebox{90}{SVHN} & \rotatebox{90}{Sun397} & \rotatebox{90}{Camelyon} & \rotatebox{90}{EuroSAT} & \rotatebox{90}{Resisc45} & \rotatebox{90}{Retinopathy} & \rotatebox{90}{Clevr-Count} & \rotatebox{90}{Clevr-Dist} & \rotatebox{90}{DMLab} & \rotatebox{90}{KITTI-Dist} & \rotatebox{90}{dSpr-Loc} & \rotatebox{90}{dSpr-Ori} & \rotatebox{90}{sNORB-Azim} & \rotatebox{90}{sNORB-Ele} & \rotatebox{90}{\textbf{Average}}\\
\midrule
\multicolumn{22}{c}{Commonly used methods} \\
\midrule
\multirow{1}{*}{Full} & 86.7 &72.2&88.0&71.4&98.3&89.5&\underline{90.1}&45.0&86.6&\textbf{96.9}&87.7&\textbf{79.4}&\underline{75.7}&59.8&54.6&78.6&79.4&\textbf{53.6}&\underline{34.6}&\underline{40.9}&\underline{74.2}\\
\multirow{1}{*}{Linear} & 0 &61.4&90.2&74.8&99.5&90.2&42.7&\textbf{55.8}&81.5&90.1&82.1&69.4&39.1&35.9&40.1&65.0&20.3&26.0&14.3&27.8&56.4\\
\multirow{1}{*}{Bias} & 0.24 &73.0&86.8&65.6&97.7&87.5&56.4&52.3&80.4&91.6&76.1&72.5&47.3&48.5&34.7&66.2&57.6&36.2&\textbf{34.7}&\textbf{66.2}&62.1\\
\midrule
\multicolumn{22}{c}{PEFT methods} \\
\midrule
\multirow{1}{*}{VPT-Deep} & 0.19 &\textbf{79.6}&90.8&\textbf{78.0}&99.5&91.4&42.3&51.7&84.9&96.2&85.0&72.0&67.6&59.4&50.1&61.3&74.4&50.6&25.7&25.7&68.6\\

\multirow{1}{*}{Adapter-8$^{\dagger}$} & 0.11 &74.2&90.7&75.9&99.5&\textbf{92.5}&38.0&\underline{55.6}&\underline{88.2}&95.0&88.9&76.9&71.4&56.7&54.1&\underline{84.5}&79.6&35.8&21.9&35.3&69.2\\

\multirow{1}{*}{LoRA$^{\dagger}$} & 0.38 &72.0&\underline{91.1}&76.5&\underline{99.6}&92.3&81.9&55.5&86.8&96.3&82.0&\underline{77.5}&71.0&\underline{60.6}&\textbf{57.0}&78.6&\textbf{87.0}&52.1&30.6&39.5&73.1\\

\multirow{1}{*}{AdaptFormer-8$^{\dagger}$} & 0.13 &74.6&91.0&76.2&\underline{99.6}&\underline{92.4}&83.5&55.4&87.0&95.2&88.9&76.8&73.8&57.9&54.8&82.7&82.2&49.7&21.6&36.9&72.6\\

\multirow{1}{*}{SSF$^{\dagger}$} & 0.27 & 75.4 & 90.0&75.1&91.0&91.2&\textbf{92.3}&55.0&\textbf{88.8}&84.2&\underline{89.1}&76.8&71.0&53.5&\underline{56.0}&83.4&84.2&\underline{52.8}&30.5&33.2&72.3\\

\midrule

\rowcolor{lightgreen}
\multirow{1}{*}{SCT (ours)} & 0.10 &\underline{75.7}&\textbf{92.6}&\underline{76.5}&\textbf{99.7}&91.7&87.2&55.5&87.6&\underline{96.5}&\textbf{89.4}&76.6&\textbf{82.5}&\textbf{63.1}&53.7&\textbf{85.9}&\underline{86.7}&46.1&26.8&40.0&\textbf{74.4}\\

\bottomrule
\end{tabular}
}
\end{center}
\vspace{-.1in}
\caption{Per-task results on VTAB-1K benchmark with a pre-trained Swin-B. The average result is calculated on all 19 datasets in the VTAB-1K benchmark. The \textbf{Bold} text represents the best performance.}
\label{tab:swin-vtab}
\end{table*}


\noindent\textbf{Performance with ViT backbone.}
We compare our SCT with the above 7 baselines in Tab.~\ref{tab:vit-vtab-full} and Fig.~\ref{fig:maingroup}. We use ViT-B/16 as the backbone and insert SCTM in each transformer layer. The default $K$ is set to 96, 1/8 of the total channels, leading to the trainable parameter number being only 0.11M.
\textbf{First}, our SCT achieves the average accuracy of 73.6\% on the 19 datasets, outperforming the full fine-tuning (85.8M) on 18 out of 19 datasets and gains the improvement of 6.3\%, 2.5\%, and 12.3\% in the three groups, respectively, with only additional 0.13\% of the backbone parameters. Such results reflect that SCT can greatly reduce the storage space and alleviate the overfitting problem common in fine-tuning large models.
\textbf{Second}, compared with Adapter (0.16M) ~\citep{adapter} that treats all the channels equally, selecting a part of salient channels for each downstream task is more effective and efficient, outperforming it by 2.3\% in average accuracy.
\textbf{Third}, compared with other PEFT methods, our SCT surpasses NOAH (0.43M) ~\citep{noah} by 0.3\% with only \textit{a quarter of} its trainable parameters. When increasing the $K$ to 192 with 0.44M extra parameters, our SCT surpasses NOAH (0.43) by 0.7 \% average accuracy. Moreover, our SCT outperforms VPT~\citep{vpt} by 3.8\%, 3.5\%, and 4.9\% in the three groups, respectively. 
These results demonstrate that instead of specially designed structures for each downstream dataset, explicitly leveraging the task-specific information can reduce the computational cost in the model search and improve the performance. Moreover, compared with SSF (0.20M)~\citep{SSF} baseline, a method without an additional trainable module, our SCT only needs almost its half parameters and achieves an average improvement of 1.0\% across 19 downstream tasks.

\noindent\textbf{Performance with Swin Transformer Backbone.}
To verify the effectiveness of SCT with different backbones, we apply SCT on hierarchical transformers, \ie, Swin-B. We use the same setting of inserting SCTM as in the ViT backbone: inserting the SCTM after the ``Attn'' block in the transformer layer. 
Considering deep layers contain more semantic information in the hierarchical structure, instead of applying SCTM on all the transformer layers, we insert it to the last half of the layers in the stage 3 and all layers of stage 4 of the Swin-B to keep a similar level of trainable parameters.
The results of Tab. \ref{tab:swin-vtab} show that SCT outperforms \textbf{full fine-tuning} in all 19 downstream tasks with only 0.12\% parameters while other methods cannot. In addition, compared with the PEFT method, SCT outperforms VPT~\citep{vpt} by 5.9\%, 3.0\%, and 7.2\% in the three groups, respectively. 
All the results above suggest that our SCT also applies to hierarchical transformers and can yield much more improvement than other PEFT methods.

\subsection{Evaluation}

\begin{table*}[ht]
\begin{center}
\large
\resizebox{\linewidth}{!}{
\begin{tabular}{lc|ccccccc|cccc|cccccccc|c}
\toprule
\multirow{2}{*}{} &  & \multicolumn{7}{c}{\textbf{Natural}} & \multicolumn{4}{c}{\textbf{Specialized}} & \multicolumn{8}{c}{\textbf{Structured}} \\
& \rotatebox{90}{\# Params (M)} & \rotatebox{90}{Cifar100} & \rotatebox{90}{Caltech101}  & \rotatebox{90}{DTD} & \rotatebox{90}{Flower102} & \rotatebox{90}{Pets} & \rotatebox{90}{SVHN} & \rotatebox{90}{Sun397} & \rotatebox{90}{Camelyon} & \rotatebox{90}{EuroSAT} & \rotatebox{90}{Resisc45} & \rotatebox{90}{Retinopathy} & \rotatebox{90}{Clevr-Count} & \rotatebox{90}{Clevr-Dist} & \rotatebox{90}{DMLab} & \rotatebox{90}{KITTI-Dist} & \rotatebox{90}{dSpr-Loc} & \rotatebox{90}{dSpr-Ori} & \rotatebox{90}{sNORB-Azim} & \rotatebox{90}{sNORB-Ele} & \rotatebox{90}{\textbf{Average}}\\
\midrule
\multirow{1}{*}{Full} & 85.8 & 68.9 & 87.7 & 64.3 & 97.2 & 86.9 & 87.4 & 38.8 & 79.7 & 95.7 & 84.2 & 73.9 &  56.3 & 58.6 & 41.7 & 65.5 & 57.5 & 46.7 & 25.7 & 29.1 & 65.6\\
\multirow{1}{*}{Linear} & 0 & 64.4 & 85.0 & 63.2 & 97.0 & 86.3 & 36.6 & 51.0 &  78.5 & 87.5 & 68.5 & 74.0 & 34.3 & 30.6 & 33.2 & 55.4 & 12.5 & 20.0 & 9.6 & 19.2 & 53.0\\
\midrule
\multirow{1}{*}{IC} & 0.11 & 68.8&91.5&71.5&99.0&91.1&89.0&54.3& 85.3&95.8&86.6&75.2& 78.1&61.2&50.3&79.0&73.1&40.0&27.4&36.7&71.1\\
\multirow{1}{*}{RC-1} & 0.11 &  72.1&90.4&71.7&99.1&91.2&90.2&54.0& 84.2&95.5&85.7&75.9& 78.8&62.5&49.5&77.6&72.2&43.3&27.3&40.9&71.7\\
\multirow{1}{*}{RC-2} & 0.11 & 74.1&91.4&72.1&99.2&90.9&90.2&55.1& 83.0&95.7&86.3&75.0& 79.3&62.5&50.4&78.2&72.5&43.2&28.1&39.8&72.0\\
\multirow{1}{*}{RC-3} & 0.11 & 71.3&91.1&70.9&99.0&90.9&89.6&54.4& 83.9&95.7&85.9&74.8& 78.6&62.2&48.9&80.2&71.2&42.1&27.8&38.5&71.4\\
\rowcolor{lightgreen}
\multirow{1}{*}{SCT} & 0.11 &75.3&91.6&72.2&99.2&91.1&91.2&55.0&85.0&96.1&86.3&76.2&81.5&65.1&51.7&80.2&75.4&46.2&33.2&45.7&73.6\\

\bottomrule
\end{tabular}
}
\end{center}
\vspace{-0.1in}
\caption{Per-task results on VTAB-1K benchmark with a pre-trained ViT-B/16. The average result is calculated on all 19 datasets in the VTAB-1K benchmark.}

\label{tab:SC-UC-Random}
\end{table*}

\begin{table*}[ht]
\begin{center}
\large
\resizebox{\linewidth}{!}{
\begin{tabular}{lc|ccccccc|cccc|cccccccc|c}
\toprule
\multirow{2}{*}{} &  & \multicolumn{7}{c}{\textbf{Natural}} & \multicolumn{4}{c}{\textbf{Specialized}} & \multicolumn{8}{c}{\textbf{Structured}} \\
& \rotatebox{90}{\# Params (M)} & \rotatebox{90}{Cifar100} & \rotatebox{90}{Caltech101}  & \rotatebox{90}{DTD} & \rotatebox{90}{Flower102} & \rotatebox{90}{Pets} & \rotatebox{90}{SVHN} & \rotatebox{90}{Sun397} & \rotatebox{90}{Camelyon} & \rotatebox{90}{EuroSAT} & \rotatebox{90}{Resisc45} & \rotatebox{90}{Retinopathy} & \rotatebox{90}{Clevr-Count} & \rotatebox{90}{Clevr-Dist} & \rotatebox{90}{DMLab} & \rotatebox{90}{KITTI-Dist} & \rotatebox{90}{dSpr-Loc} & \rotatebox{90}{dSpr-Ori} & \rotatebox{90}{sNORB-Azim} & \rotatebox{90}{sNORB-Ele} & \rotatebox{90}{\textbf{Average}}\\
\midrule
\multirow{1}{*}{Full} & 85.8 & 68.9 & 87.7 & 64.3 & 97.2 & 86.9 & 87.4 & 38.8 & 79.7 & 95.7 & 84.2 & 73.9 &  56.3 & 58.6 & 41.7 & 65.5 & 57.5 & 46.7 & 25.7 & 29.1 & 65.6\\
\multirow{1}{*}{Linear} & 0 & 64.4 & 85.0 & 63.2 & 97.0 & 86.3 & 36.6 & 51.0 &  78.5 & 87.5 & 68.5 & 74.0 & 34.3 & 30.6 & 33.2 & 55.4 & 12.5 & 20.0 & 9.6 & 19.2 & 53.0\\
\midrule
\multirow{1}{*}{w/o CA} & 0.11 &75.1&91.8&71.6&99.1&90.9&91.2&55.1&84.4&95.6&85.6&75.1& 81.1&64.7&51.2&78.3&75.0&44.8&31.3&40.4&72.8\\

\multirow{1}{*}{w/ CA} & 0.11 &75.3&91.6&72.2&99.2&91.1&91.2&55.0&85.0&96.1&86.3&76.2&81.5&65.1&51.7&80.2&75.4&46.2&33.2&45.7&73.6\\

\bottomrule
\end{tabular}
}
\end{center}
\vspace{-.1in}
\caption{Per-task results of evaluating the effectiveness of adopting the Class-Aware Importance Score. The average result is calculated on all 19 datasets in the VTAB-1K benchmark.}
\label{tab:cacompare}
\end{table*}

\noindent\textbf{Effectiveness of Salient Channel Tuning.}
We compare three channel selection strategies to verify the effectiveness of selecting salient channels in Tab.~\ref{tab:SC-UC-Random}. The selection strategies include Salient Channel Selection (SC), Inconspicuous Channel Selection (IC), and Random Channel Selection (RC).
IC selects channels with the lowest $K$ values of $L_2$ norm, and RC denotes selecting $K$ channels randomly.
We randomly select three sets of channels (RC-1/2/3) for random channel selection to alleviate the outliers. As shown in Tab.~\ref{tab:SC-UC-Random}, SC achieves the best results and outperforms IC by 2.4\% in the average accuracy. Random channel selection could obtain better results than full fine-tuning, but all of them perform worse than SC.
Interestingly, even IC can perform better than full fine-tuning, demonstrating that selecting a small subset of the channels can prevent the large model from overfitting to the small training set.

\noindent\textbf{Effectiveness of Class-Aware Calculation for importance score.} In Sec.~\ref{sec:class-aware}, we introduced our method, which considers the impact of individual classes rather than treating the entire dataset as a whole when estimating the importance score (IS). This approach aims to identify the most significant channels based on the mutual information between categories and reduce the effects of imbalanced data. Our experimental results, presented in Tab.~\ref{tab:cacompare}, demonstrate that adopting a class-aware calculation approach improves performance across all three groups.

\begin{table*}[ht]
\begin{center}
\large
\resizebox{\linewidth}{!}{
\begin{tabular}{lc|ccccccc|cccc|cccccccc|c}
\toprule
\multirow{2}{*}{} &  & \multicolumn{7}{c}{\textbf{Natural}} & \multicolumn{4}{c}{\textbf{Specialized}} & \multicolumn{8}{c}{\textbf{Structured}} \\
& \rotatebox{90}{\# Params (M)} & \rotatebox{90}{Cifar100} & \rotatebox{90}{Caltech101}  & \rotatebox{90}{DTD} & \rotatebox{90}{Flower102} & \rotatebox{90}{Pets} & \rotatebox{90}{SVHN} & \rotatebox{90}{Sun397} & \rotatebox{90}{Camelyon} & \rotatebox{90}{EuroSAT} & \rotatebox{90}{Resisc45} & \rotatebox{90}{Retinopathy} & \rotatebox{90}{Clevr-Count} & \rotatebox{90}{Clevr-Dist} & \rotatebox{90}{DMLab} & \rotatebox{90}{KITTI-Dist} & \rotatebox{90}{dSpr-Loc} & \rotatebox{90}{dSpr-Ori} & \rotatebox{90}{sNORB-Azim} & \rotatebox{90}{sNORB-Ele} & \rotatebox{90}{\textbf{Average}}\\
\midrule
\multirow{1}{*}{Full} & 85.8 & 68.9 & 87.7 & 64.3 & 97.2 & 86.9 & 87.4 & 38.8 & 79.7 & 95.7 & 84.2 & 73.9 &  56.3 & 58.6 & 41.7 & 65.5 & 57.5 & 46.7 & 25.7 & 29.1 & 65.6\\
\multirow{1}{*}{Linear} & 0 & 64.4 & 85.0 & 63.2 & 97.0 & 86.3 & 36.6 & 51.0 &  78.5 & 87.5 & 68.5 & 74.0 & 34.3 & 30.6 & 33.2 & 55.4 & 12.5 & 20.0 & 9.6 & 19.2 & 53.0\\
\midrule
\multirow{1}{*}{SCT$_\text{MLP}$} & 0.11 & 72.9&92.0&71.6&99.1&90.6&90.5&54.9& 86.2&95.9&85.9&76.0& 81.3&64.7&50.7&78.9&77.2&43.3&29.3&43.6&72.9\\

\multirow{1}{*}{SCT$_\text{Attn}$} & 0.11 &75.3&91.6&72.2&99.2&91.1&91.2&55.0&85.0&96.1&86.3&76.2&81.5&65.1&51.7&80.2&75.4&46.2&33.2&45.7&73.6\\

\bottomrule
\end{tabular}
}
\end{center}
\vspace{-.1in}
\caption{Per-task results of evaluating the effectiveness of the insert positions. The average result is calculated on all 19 datasets in the VTAB-1K benchmark.}

\label{tab:insertcompare}
\end{table*}

\begin{table*}[ht]
\begin{center}
\large
\resizebox{\linewidth}{!}{
\begin{tabular}{lc|ccccccc|cccc|cccccccc|c}
\toprule
\multirow{2}{*}{} &  & \multicolumn{7}{c}{\textbf{Natural}} & \multicolumn{4}{c}{\textbf{Specialized}} & \multicolumn{8}{c}{\textbf{Structured}} \\
& \rotatebox{90}{\# Params (M)} & \rotatebox{90}{Cifar100} & \rotatebox{90}{Caltech101}  & \rotatebox{90}{DTD} & \rotatebox{90}{Flower102} & \rotatebox{90}{Pets} & \rotatebox{90}{SVHN} & \rotatebox{90}{Sun397} & \rotatebox{90}{Camelyon} & \rotatebox{90}{EuroSAT} & \rotatebox{90}{Resisc45} & \rotatebox{90}{Retinopathy} & \rotatebox{90}{Clevr-Count} & \rotatebox{90}{Clevr-Dist} & \rotatebox{90}{DMLab} & \rotatebox{90}{KITTI-Dist} & \rotatebox{90}{dSpr-Loc} & \rotatebox{90}{dSpr-Ori} & \rotatebox{90}{sNORB-Azim} & \rotatebox{90}{sNORB-Ele} & \rotatebox{90}{\textbf{Average}}\\
\midrule
\multirow{1}{*}{Full} & 85.8 & 68.9 & 87.7 & 64.3 & 97.2 & 86.9 & 87.4 & 38.8 & 79.7 & 95.7 & 84.2 & 73.9 &  56.3 & 58.6 & 41.7 & 65.5 & 57.5 & 46.7 & 25.7 & 29.1 & 65.6\\
\multirow{1}{*}{Linear} & 0 & 64.4 & 85.0 & 63.2 & 97.0 & 86.3 & 36.6 & 51.0 &  78.5 & 87.5 & 68.5 & 74.0 & 34.3 & 30.6 & 33.2 & 55.4 & 12.5 & 20.0 & 9.6 & 19.2 & 53.0\\
\multirow{1}{*}{Bias} & 0.10 & 72.8&87.0&59.2&97.5&85.3&59.9&51.4&78.7&91.6&72.9&69.8&61.5&55.6&32.4&55.9&66.6&40.0&15.7&25.1&62.1\\
\midrule
\multirow{1}{*}{SCT$_\text{K=32}$} & 0.01 &71.7&90.3&70.2&99.0&90.1&85.8&53.1&82.2&95.6&84.8&75.7&71.6&62.1&48.3&74.0&59.1&42.0&24.5&33.6&69.1\\

\multirow{1}{*}{SCT$_\text{K=96}$} & 0.11 &75.3&91.6&72.2&99.2&91.1&91.2&55.0&85.0&96.1&86.3&76.2&81.5&65.1&51.7&80.2&75.4&46.2&33.2&45.7&73.6\\

\multirow{1}{*}{SCT$_\text{K=192}$} & 0.44 &76.0&91.5&72.3&99.1&91.2&92.2&55.2&86.2&95.7&87.0&76.1&82.5&64.0&52.1&81.0&75.5&47.5&34.7&45.6&74.0\\

\bottomrule
\end{tabular}
}
\end{center}
\vspace{-.1in}
\caption{Per-task results of evaluating different $K$ values on VTAB-1K benchmark.}
\label{tab:vit-vtab-Kvalue-full}
\end{table*}

\begin{figure*}[htbp]
\begin{center}
\includegraphics[width=0.48\linewidth]{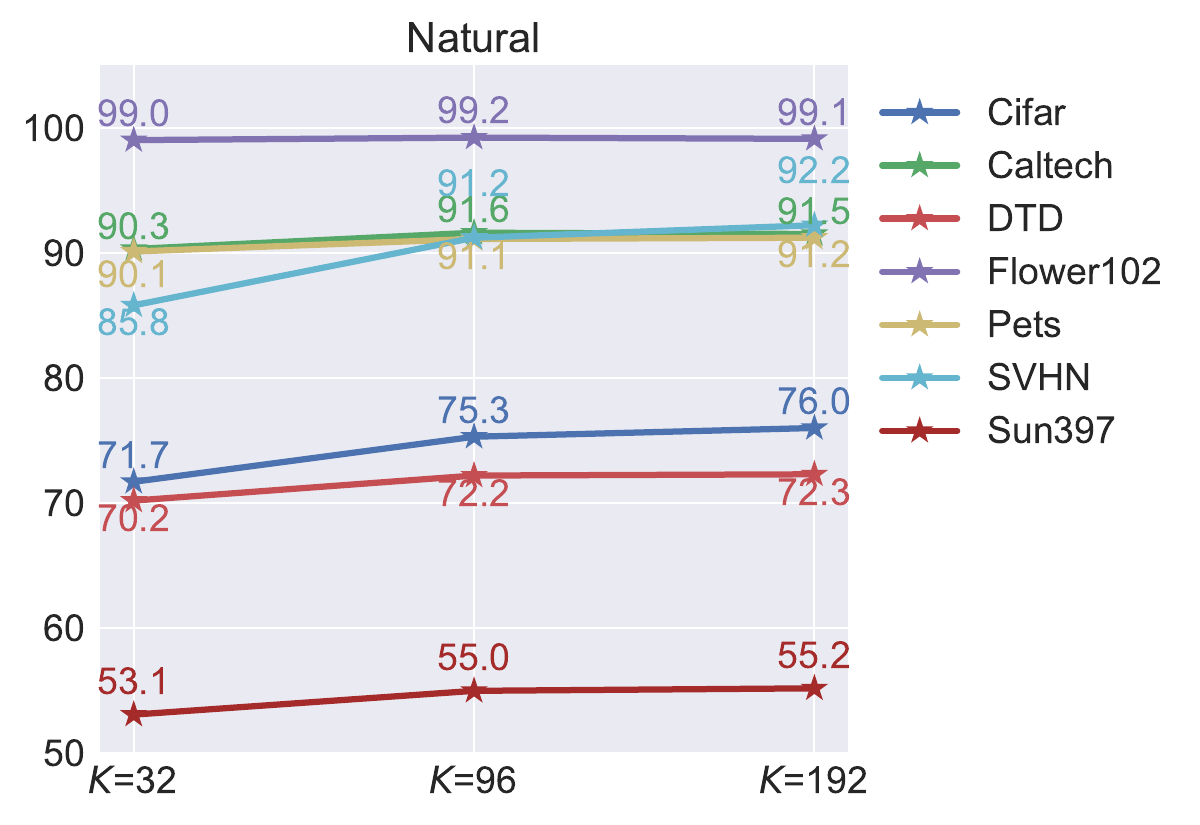}
\includegraphics[width=0.48\linewidth]{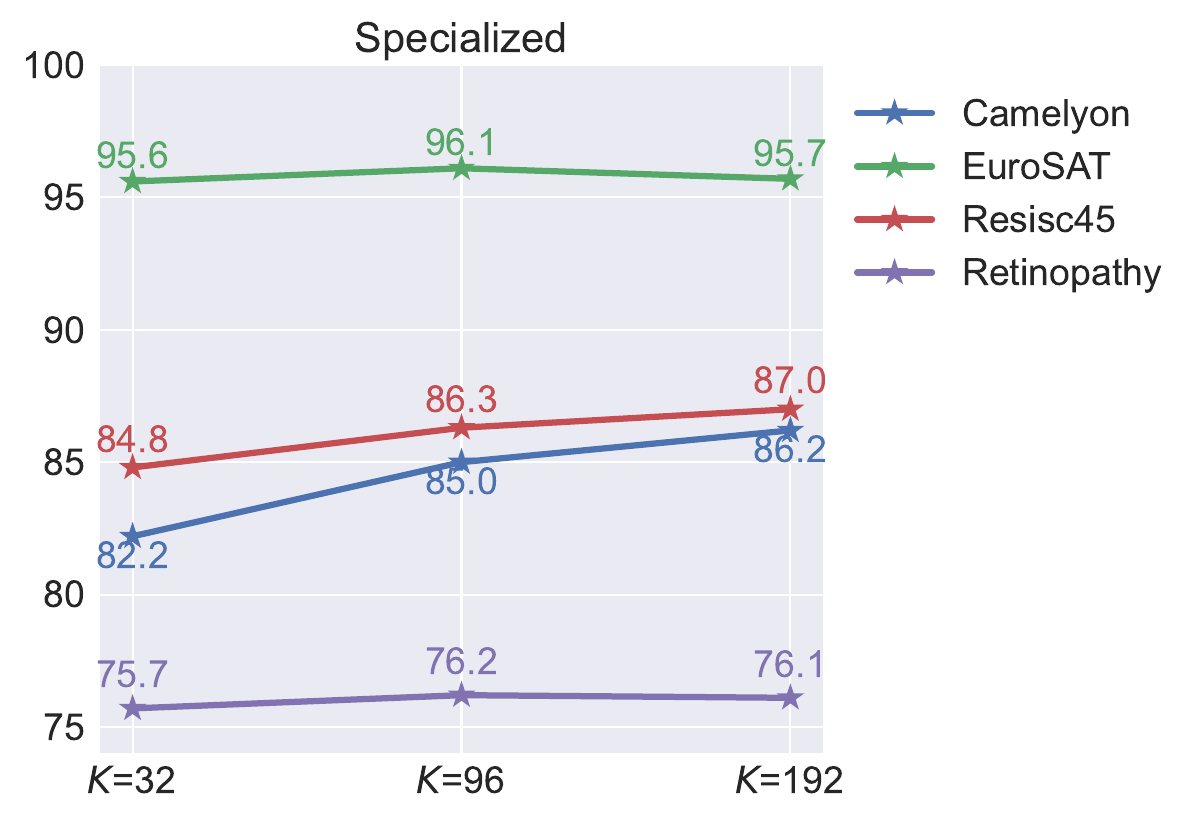}
\includegraphics[width=0.48\linewidth]{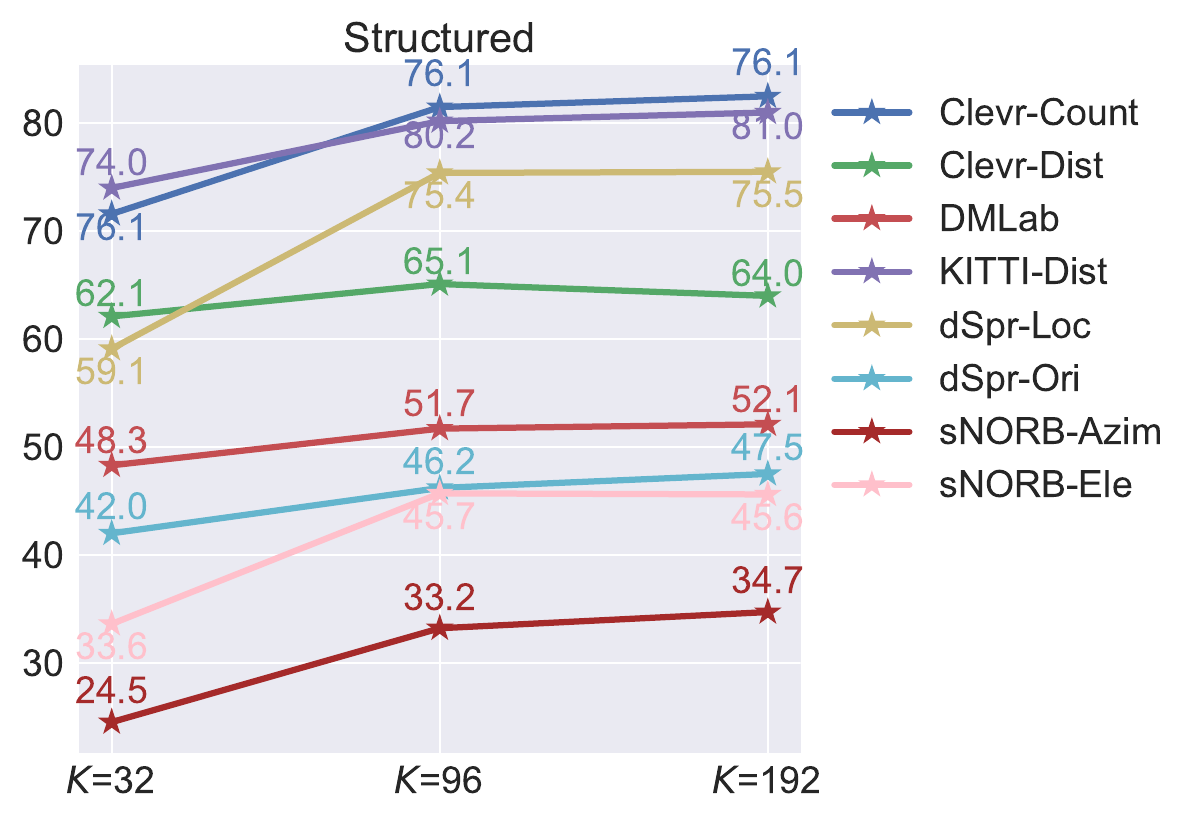}
\end{center}
\caption{Choosing different $K$ values on 19 downstream tasks.}
\label{fig:comk}
\end{figure*}

\begin{table}[]
\begin{center}
\resizebox{\linewidth}{!}{
\begin{tabular}{lcccc}
\toprule
 & Cifar100 & Resisc45 & DMLab & Params\\
\midrule
\multirow{1}{*}{SCT$_\text{Attn}$}  & 75.3 & 86.3 & \textbf{51.7} & 0.11M \\
\multirow{1}{*}{SCT$_\text{MLP}$}  & 72.9 & 85.9 & 50.9 & 0.11M\\
\multirow{1}{*}{SCT$_\text{Attn}$ + SCT$_\text{MLP}$}  &\textbf{75.7} & \textbf{86.5} & 51.0 & 0.22M \\
\bottomrule
\end{tabular}
}
\end{center}
\caption{Comparison of different insert positions. The \textbf{Bold} text represents the best performance.}
\label{tab:insertcompare2}
\end{table}

\noindent\textbf{Insert Position.}
As shown in Fig.~\ref{fig:insertposition}, SCTM can be inserted after the MLP block (SCT$_\text{MLP}$) or between the MHSA block and MLP block (SCT$_\text{Attn}$). To investigate the influence of insert location, we compare two forms on the VTAB-1K benchmark in Tab.~\ref{tab:insertcompare}. Both achieve promising performances, and SCT$_\text{Attn}$ outperforms SCT$_\text{MLP}$ on two of the three groups. 
We conjecture that after gathering long-range dependencies with MHSA, the salient channels of features are more representative, which can be better adapted to downstream tasks. We also evaluate the performance of inserting SCTM after both MHSA and MLP blocks, as shown in Tab.~\ref{tab:insertcompare2}. Jointly inserting SCT$_\text{Attn}$ and SCT$_\text{MLP}$ only brings small improvements while the number of parameters doubles.

\begin{figure*}[t]
\centering
\includegraphics[width=0.9\linewidth]{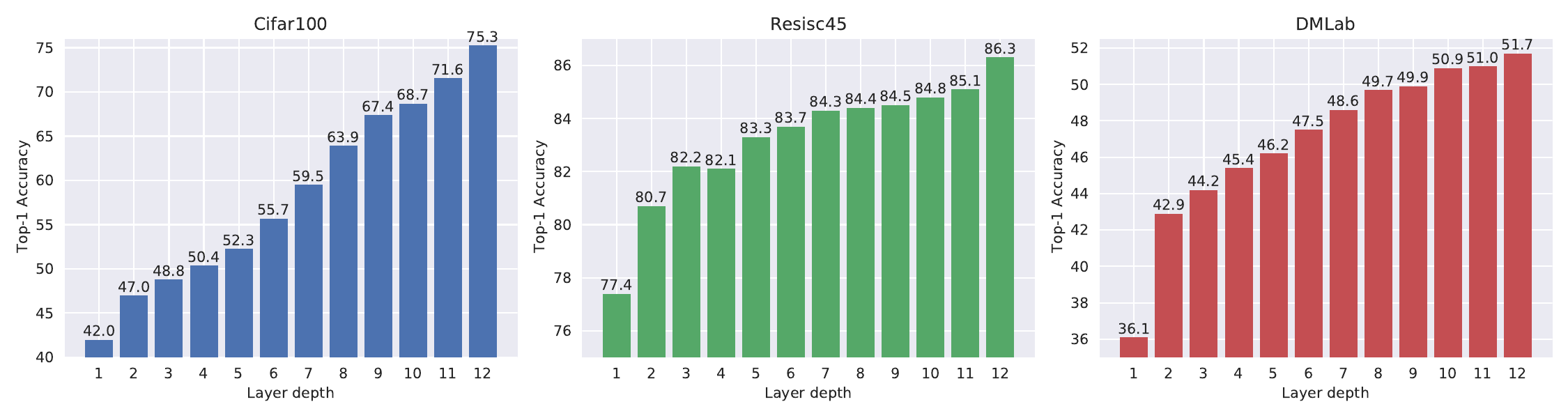}
\vspace{-0.15in}
\captionsetup{font=scriptsize}
\caption{Results of inserting SCTM into different layers. Zoom in for the best view.}
\label{fig:layerdepth}
\end{figure*}

\noindent\textbf{Number of insert depth.}
The insert depth is an important factor that can significantly influence the performance of a model. To investigate this, we conducted experiments by inserting SCTM into the last $l$ layers of ViT-B and evaluated the performance on three representative tasks from each group of VTAB-1K. Our findings, presented in Fig.~\ref{fig:layerdepth}, reveal that increasing the insert depth results in a gradual improvement in performance. For the Resisc45 and DMLab tasks, we observed that competitive performance was achieved when the SCTM was inserted into the last six layers. Further increases in the insert depth resulted in a plateau in performance, indicating that additional layers did not contribute significantly to the final performance.

\noindent\textbf{Number of selected channels $K$.}
The most important hyperparameter of SCTM is the number of selected channels $K$, which influences the model architecture and the number of trainable parameters. \textbf{Note that} different from previous works~\citep{vpt,noah} that select hyperparameters for each dataset, we use the same $K$ for all the datasets. As shown in Tab.\ref{tab:vit-vtab-Kvalue-full}, SCT$_\text{K=32}$ beats the full fine-tuning and bias tuning by 3.5\% and 7.0\%, respectively. Furthermore, SCT$_\text{K=32}$ only adopts 0.01M parameters while the tuning bias term adopts 0.10M. As shown in Fig.~\ref{fig:comk}, the performance generally improves along with the increase of $K$. However, the improvement of $K=192$ over $K=96$ is marginal, while the number of parameters is four times larger. Considering both the effectiveness and efficiency, we set $K$ to 96 by default.

\subsection{Computation Analysis}
Our analysis considers a ViT-B backbone with $L$ layers and $D$ dimensions and $N$ tokens for a single image. We also assume that the intermediate dimension of Adapter~\citep{adapter} is $D^{\prime}$, that the prompt length of VPT~\citep{vpt} is $n$, and that the total insert times of SSF~\citep{SSF} is $m$. Finally, we compare our proposed SCT approach to Adapter, VPT, and SSF in terms of parameters and FLOPs, as summarized in  Tab.~\ref{tab:complexity}. Notably, our selection of $K$ as $\frac{1}{8}D$ is fairly small compared to $D$. When the reduction ratio of the Adapter is also defined as 8, the parameters of the Adapter are $\frac{1}{4}LD^2$ while our parameters are $\frac{1}{64}LD^2$. When we compare our approach to SSF, which inserts its module $m$ times in the ViT-B backbone, the number of parameters for SSF is $mLD$. Examining the ViT-B backbone, we find that $m=74$ and $\frac{1}{64}D=12$, our parameters ($12LD < 74LD$) and FLOPs ($12NLD < 74NLD$) are smaller than SSF. Overall, our analysis suggests that SCT may offer PEFT a more efficient and effective baseline.

\begin{figure*}[!h]
\begin{center}
\includegraphics[width=0.24\linewidth]{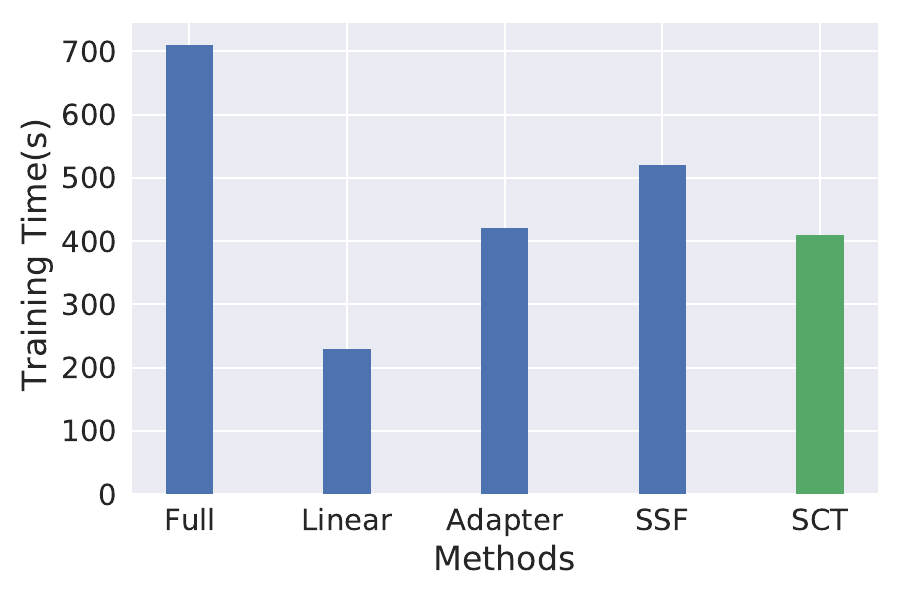}
\includegraphics[width=0.24\linewidth]{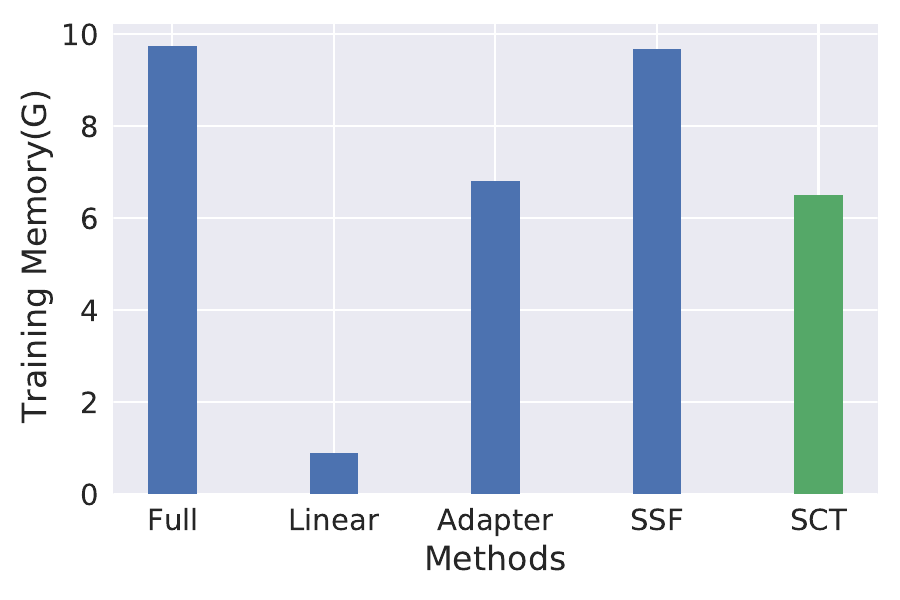}
\includegraphics[width=0.24\linewidth]{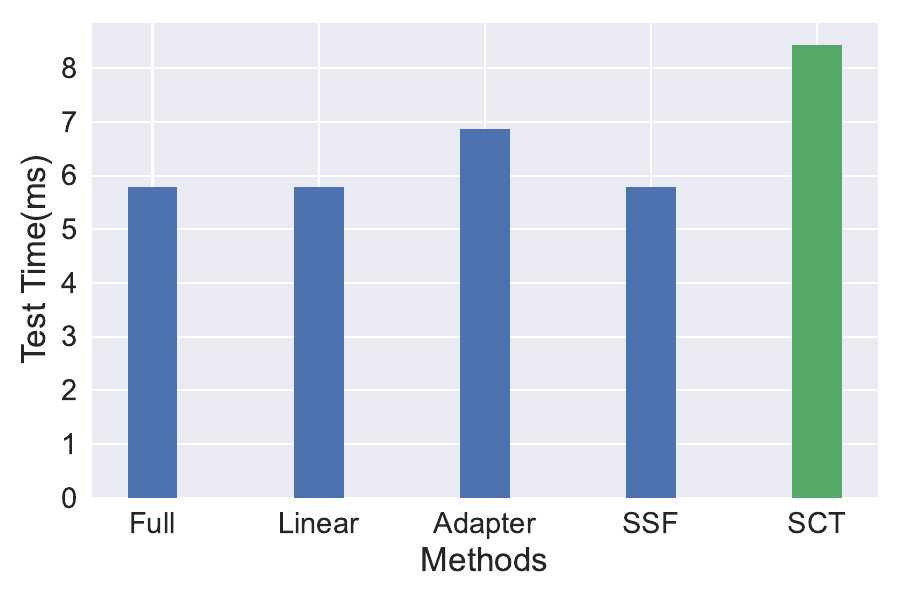}
\includegraphics[width=0.24\linewidth]{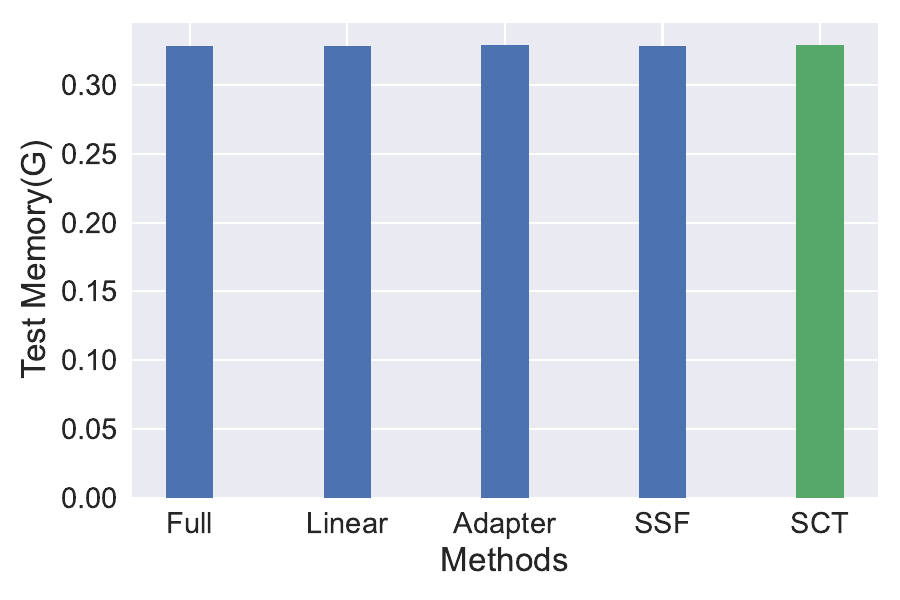}
\end{center}
 \vspace{-0.1in}
\caption{The computational costs of training and test phases.}
\label{fig:computation}
\end{figure*}

To assess the computational efficiency of our SCT approach, we choose to compare it with Adapter (0.16M) and SSF (0.20M) since they have a similar level of parameters and all of them are applied in the backbone network. We measured the running time and memory usage for both the training and testing phases using a batch size of 64 on the Cifar100 (VTAB-1K) dataset with the ViT-B backbone. We conducted training experiments on a single NVIDIA V100-32GB GPU and test experiments on a single NVIDIA A100-40GB GPU. As the usage of GPU memory during the training and testing phases is different. During the training phase, tensor activations, model parameters, gradients, and optimizer states (model, gradient, momentum) are the main sources. During the testing phase, tensor activations and model parameters are the main sources. Our results, as shown in Fig.~\ref{fig:computation}, indicate that our SCT approach outperforms Adapter and SSF regarding training time and memory usage. As our tunable parameters are smaller than Adapter and SSF, the cost of gradients and optimizer states are small which helps us save the GPU memory during training. Specifically, our method requires only 12 additional SCT computations, while SSF requires 74 scaling and shifting operations. The training of SSF is not quite efficient compared with our SCT. However, during testing, the test time of our method is slightly higher than Adapter and SSF. Noted, SSF could merge its extra parameters to the original backbone with no extra inference costs while our method does not focus on reducing the computation costs during inference. The reason for higher test costs compared with Adapter may be the extra operations of querying the salient channels from a full feature and then putting the projected salient channels back. Furthermore, we give exact test time cost results in Tab. \ref{tab:runtime}. Here LoRA and SSF could merge its parameter into the backbone network thus we choose Full fine-tuning as the representative.

\begin{table}[th]
\begin{center}
\resizebox{\linewidth}{!}{
\begin{tabular}{lcccc}
\toprule
 Methods & Test time & Memory & Params\\
\midrule
\multirow{1}{*}{Full fine-tuning}  & 5.78 ms & 336 MB& 85.8 M&  \\

\multirow{1}{*}{Adapter-8}  & 6.87 ms & 337 MB & 85.8+0.16 M &  \\

\multirow{1}{*}{AdaptFormer-8}  & 7.88 ms & 337 MB & 85.8+0.18 M& \\

\multirow{1}{*}{SCT (Ours)}  & 8.43 ms & 337 MB & 85.8+0.11 M&  \\

\bottomrule
\end{tabular}
}
\end{center}
\vspace{-0.1in}
\caption{We evaluate the test time for 1 image with resolution 224. We run 200 times and yield the average result.}
\vspace{-0.2in}
\label{tab:runtime}
\end{table}

\begin{table}[t]
\small
\centering
\resizebox{\linewidth}{!}{
\begin{tabular}{l|cccc}
\toprule
& Adapter & VPT-Deep & SSF  & TTC-Module \\
\midrule
\multirow{1}{*}{\# Extra Parameters} & $2LDD^{\prime}$& $nLD$ & $mLD$ & $LKK$ \\
\multirow{1}{*}{\# Extra FLOPs} & $2NLDD^{\prime}$ & $2n(2N+n)LD$ & $mNLD$ & $NLKK$  \\
\bottomrule
\end{tabular}
}
\caption{A complexity analysis of Adapter~\citep{adapter}, VPT~\citep{vpt}, SSF~\citep{SSF}, and our proposed TTC-Module.}
\label{tab:complexity}
\vspace{-0.2in}
\end{table}

\subsection{Experiments on Domain Generalization}

\begin{table*}[]
\small
\centering
\resizebox{0.7\linewidth}{!}{
\begin{tabular}{l|c|cccc}
\toprule
\multirow{2}{4em}{} & {\textbf{Source}} & \multicolumn{4}{c}{\textbf{Target}} \\
& ImageNet & -V2 & -Sketch  & -A & -R \\
\midrule
\multirow{1}{*}{Adapter~\citep{adapter}}  & 70.5 & 59.1 & 16.4 & 5.5 & 22.1 \\
\multirow{1}{*}{VPT~\citep{vpt}}  & 70.5 & 58.0 & 18.3 & 4.6 & 23.2 \\
\multirow{1}{*}{LoRA~\citep{lora}}  & 70.8 & 59.3 & 20.0 & 6.9 & 23.3 \\
\multirow{1}{*}{NOAH~\citep{noah}} & 71.5 & \textbf{66.1} & 24.8 & 11.9 & 28.5 \\
\midrule
\multirow{1}{*}{SCT-B (ours)} & \textbf{77.1} & 65.8 & \textbf{28.5} & \textbf{12.1} & \textbf{31.0} \\
\bottomrule
\end{tabular}
}
\caption{Comparison with previous methods on domain generalization. The backbone network is ViT-B/16. The number of salient channels is 192. The \textbf{Bold} text represents the best performance.}
\label{tab:dg}
\end{table*}

\noindent\textbf{Dataset.}
In addition to evaluating the model on test data of the same distribution, modern deep neural networks commonly suffer from performance degradation when the testing distribution is different from that of the training set, \ie, domain shift, which is inevitable in a real-world application. To alleviate this problem, domain generalization~\citep{zhou2021mixstyle,zhao2022shade,zhou2022device,yang2022openood,yang2022fsood,yang2021oodsurvey} is investigated in the community, aiming to train a model with one or multiple source domains that can perform well on other unseen target domains. To verify the generalization ability of our SCT, we follow \cite{noah} to conduct experiments on ImageNet and its variants.
Specifically, we use the ImageNet-1K~\citep{deng2009imagenet} as the source domain with 16-shot per category and evaluate our model on ImageNetV2~\citep{recht2019imagenet}, ImageNet-Sketch~\citep{wang2019learning}, ImageNet-A~\citep{hendrycks2021natural}, and ImageNet-R~\citep{hendrycks2021many}.
ImageNetV2~\citep{recht2019imagenet} is collected from different sources from ImageNet-1K with the same protocol, and ImageNet-Sketch~\citep{wang2019learning} contains the sketch images of ImageNet classes. Both of them use the same classes as ImageNet-1K. ImageNet-A~\citep{hendrycks2021natural} and ImageNet-R~\citep{hendrycks2021many} contain the adversarially-filtered images and renditions of ImageNet data of a 200-class subset, respectively. We use a large version of SCT, \ie, SCT-B, containing comparable parameters with NOAH (0.44M \vs 0.43M).


\noindent\textbf{Results.}
In Tab.~\ref{tab:dg}, we compare our SCT-B with Adapter~\citep{adapter}, VPT~\citep{vpt}, LoRA~\citep{lora}, and NOAH~\citep{noah} on the above datasets to verify the generalization ability. We can make two observations. \textbf{First}, SCT-B outperforms the previous best method~(NOAH) on three of the four target datasets and achieves comparable performance on ImageNetV2. Specifically, SCT-B yields an improvement of 2.5\% on ImageNet-R over NOAH. \textbf{Second}, our SCT-B achieves an accuracy of 77.1\% on the source domain, greatly outperforming previous methods by 6\%. Since the backbone model is pre-trained on ImageNet-21K, the results on ImageNet-1K show that SCT can better enhance the knowledge transfer from superset to subset. The two observations demonstrate the superiority of our SCT over previous fine-tuning techniques on strong generalization ability.

\subsection{Experiments on Few-shot Learning}

\begin{figure*}[!htp]
\begin{center}
\includegraphics[width=0.31\linewidth]{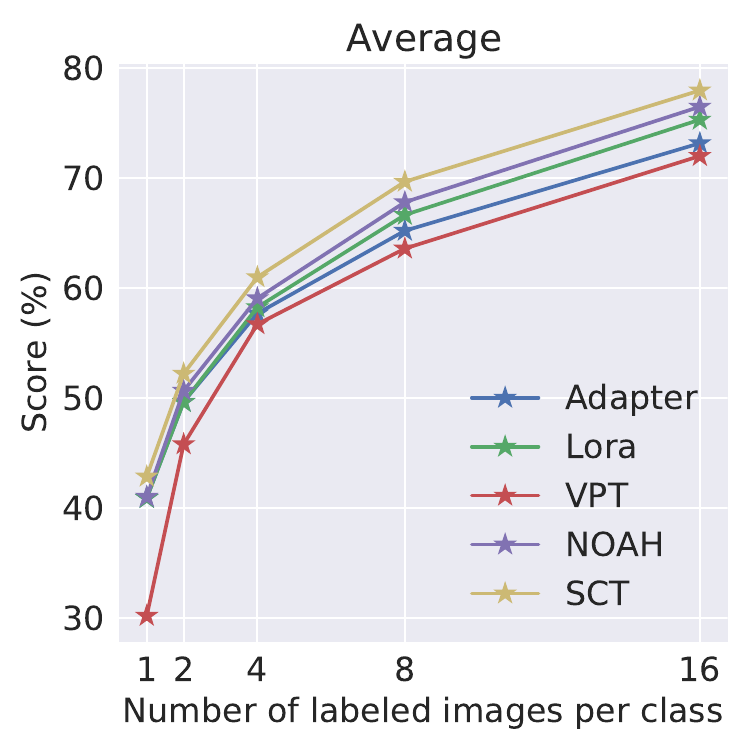}
\includegraphics[width=0.31\linewidth]{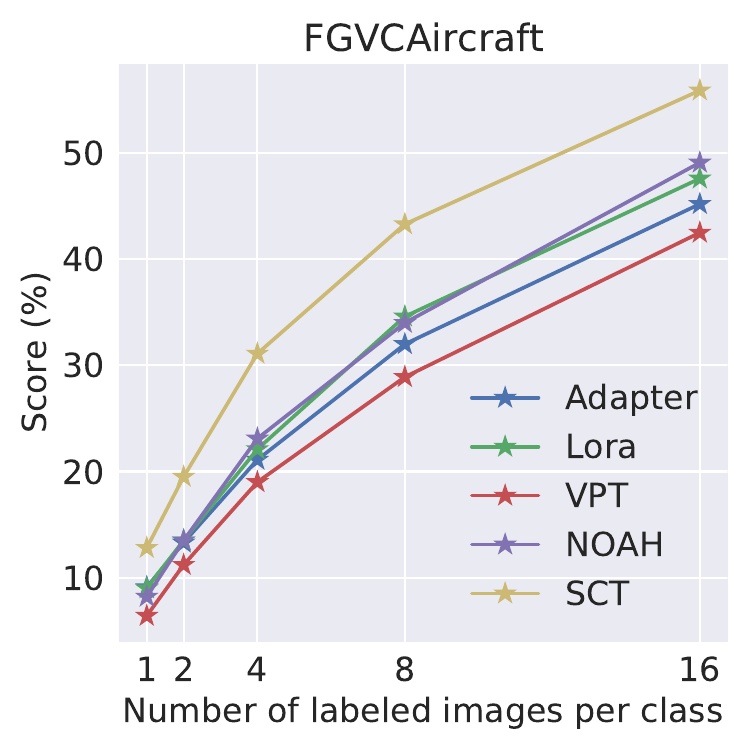}
\includegraphics[width=0.31\linewidth]{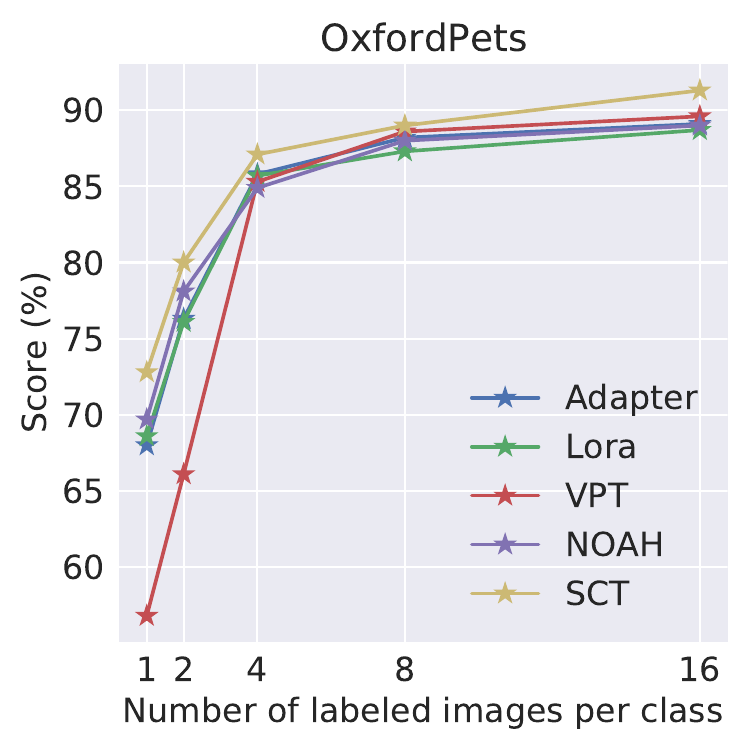}
\includegraphics[width=0.31\linewidth]{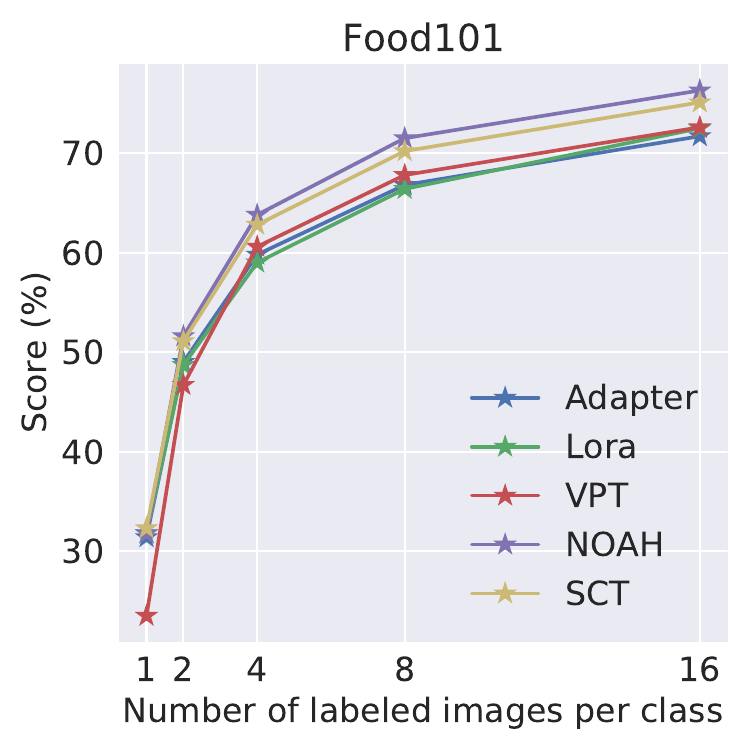}
\includegraphics[width=0.31\linewidth]{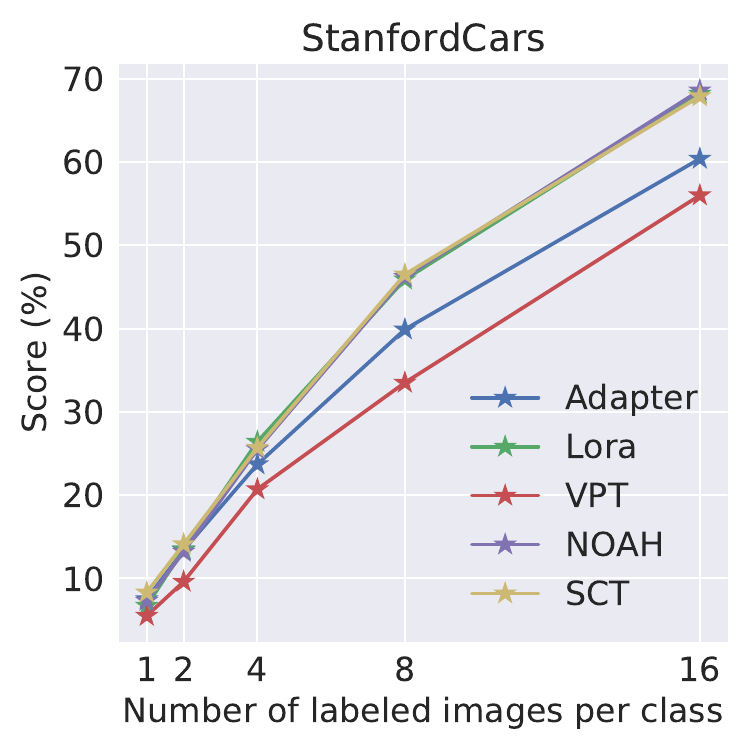}
\includegraphics[width=0.31\linewidth]{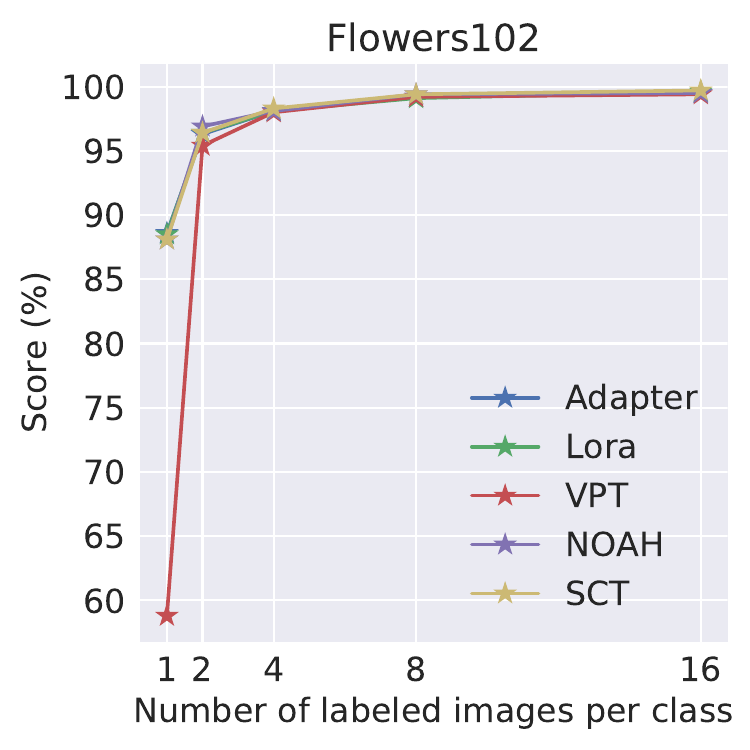}
\end{center}
\caption{Results of few-shot learning on five fine-grained visual recognition datasets.}
\label{fig:fewshot}
\end{figure*}

\noindent\textbf{Datasets.}
Following \cite{noah}, we choose five fine-grained visual recognition datasets, including Food101 \citep{bossard2014food}, Flowers102 \citep{nilsback2006visual}, StandfordCars \citep{krause20133d}, OxfordPets \citep{parkhi2012cats}, and FGVCAircraft \citep{maji2013fine}, to investigate the effectiveness of SCT in few-shot learning. The categories in these datasets cover various visual concepts closely related to our daily life: food, plant, vehicle, and animal.
Next, we follow existing studies \citep{noah, jie2022convolutional} to evaluate our model under 1, 2, 4, 8, and 16 shots settings. The experimental setup is the same as in VTAB-1K.

\noindent\textbf{Results.} 
According to Fig. \ref{fig:fewshot}, our SCT surpasses other baselines in all settings in terms of average performance. In addition, SCT largely obtains the advantages at FGVCAircraft and OxfordPets datasets. For other datasets, SCT achieves competitive performance compared with other baselines. Such results demonstrate that SCT has a strong generalization ability that can be readily transferred to downstream tasks only with a few samples.

\section{Conclusion}
This paper introduces a simple baseline for PEFT, which is both simple and effective. Unlike previous methods, we adopt a channel selection perspective and propose a straightforward importance score to identify task-specific channels. By eliminating downsampling and nonlinearity operations, our method outperforms the similar Adapter method, requiring fewer parameters. We evaluate our algorithm on 19 downstream tasks and achieve the best results with only 780$\times$ fewer parameters than the original backbone. We also test our approach on four datasets with natural distribution shifts to assess domain generalization capabilities and on five datasets in the few-shot scenario to evaluate performance in low-data regimes. Our approach is computationally efficient during both training and testing.
Moreover, it is independent to prompt-based and LoRa-like methods, allowing for further exploration in future research. Although our channel selection technique is straightforward, there is considerable potential to enhance its performance. Our baseline can be particularly useful in resource-limited situations as it requires fewer extra parameters and enables rapid adaptation to new tasks with just a few samples, making the simple and small modules effective knowledge storage units.

\section*{Acknowledgement}

This research is supported by National Research Foundation, Singapore and A*STAR, under its RIE2020 Industry Alignment Fund – Industry Collaboration Projects (IAF-ICP) grant call (Grant No. I2001E0059) – SIA-NUS Digital Aviation Corp Lab. Mike Zheng Shou is supported by the National Research Foundation, Singapore, under its NRFF Award NRF-NRFF13-2021-0008, and Mike Zheng Shou's Start-Up Grant from NUS.














\bibliographystyle{spbasic}      
\bibliography{ref}   

\end{document}